\def\year{2022}\relax
\documentclass[letterpaper]{article} 
\usepackage{aaai22}  
\usepackage{times}  
\usepackage{helvet}  
\usepackage{courier}  
\usepackage[hyphens]{url}  
\usepackage{graphicx} 
\usepackage{natbib}  
\usepackage{caption} 
\DeclareCaptionStyle{ruled}{labelfont=normalfont,labelsep=colon,strut=off} 
\usepackage{algorithm}
\usepackage{algorithmic}
\newcommand{\figref}[1]{Figure \ref{#1}}
\newcommand{\tabref}[1]{Table \ref{#1}}

\usepackage{newfloat}
\usepackage{listings}
\floatstyle{ruled}
\newfloat{listing}{tb}{lst}{}
\floatname{listing}{Listing}

\usepackage{latexsym}
\usepackage{adjustbox}
\usepackage{natbib}
\usepackage{booktabs}
\usepackage{tablefootnote}
\usepackage[flushleft]{threeparttable} 
\usepackage{booktabs,caption}
\usepackage{multirow}
\usepackage{sidecap}
\usepackage{graphics}
\usepackage{graphicx} 
\usepackage[subrefformat=parens]{subcaption}
\usepackage{xcolor}
\usepackage{caption}
\usepackage{tablefootnote}
\usepackage{amsmath}
\usepackage{caption,stackengine}
\usepackage{parselines} 
\usepackage{xspace} 
\usepackage{footnote}

\usepackage{nccmath}
\usepackage{breakcites}
\usepackage{cite}
\usepackage{pifont}
\usepackage{kotex}
\usepackage{tabularx}
\usepackage{amsmath}
\usepackage{newtxtext,newtxmath}
\usepackage{array}
\usepackage{makecell}
\usepackage{color}
\usepackage{colortbl}
\usepackage{footnote}
\usepackage{verbatim}
\usepackage{stackengine}
\title{Call for Customized Conversation: \\ Customized Conversation Grounding Persona and Knowledge}

\author{Yoonna Jang\textsuperscript{\rm 1}\equalcontrib, Jungwoo Lim\textsuperscript{\rm 1}\equalcontrib, Yuna Hur\textsuperscript{\rm 1}\equalcontrib, Dongsuk Oh\textsuperscript{\rm 1}, Suhyune Son\textsuperscript{\rm 1}, \\ Yeonsoo Lee\textsuperscript{\rm 2}, Donghoon Shin\textsuperscript{\rm 2}, Seungryong Kim\textsuperscript{\rm 1}\footnote{These authors are the corresponding authors.}, and Heuiseok Lim\textsuperscript{\rm 1}$^{\dagger}$}

\affiliations{
\textsuperscript{\rm 1}Department of Computer Science and Engineering, Korea University \\ 
\textsuperscript{\rm 2}Language AI Lab, NCSOFT \\ 
\{morelychee, wjddn803, yj72722, inow3555, ssh5131, seungryong\_kim, limhseok\}@korea.ac.kr \\
\{yeonsoo, dhshin\}@ncsoft.com
}

\begin{document}

\maketitle
\begin{abstract}
Humans usually have conversations by making use of prior knowledge about a topic and background information of the people whom they are talking to. However, existing conversational agents and datasets do not consider such comprehensive information, and thus they have a limitation in generating the utterances where the knowledge and persona are fused properly. To address this issue, we introduce a \textit{call For Customized conversation} (FoCus) dataset where the customized answers are built with the user's persona and Wikipedia knowledge. To evaluate the abilities to make informative and customized utterances of pre-trained language models, we utilize BART and GPT-2 as well as transformer-based models. We assess their generation abilities with automatic scores and conduct human evaluations for qualitative results. We examine whether the model reflects adequate persona and knowledge with our proposed two sub-tasks, persona grounding (PG) and knowledge grounding (KG). Moreover, we show that the utterances of our data are constructed with the proper knowledge and persona through grounding quality assessment.
\end{abstract}

\section{Introduction}
\begin{figure}[t]
	\centering
	\includegraphics[width=0.9\linewidth]{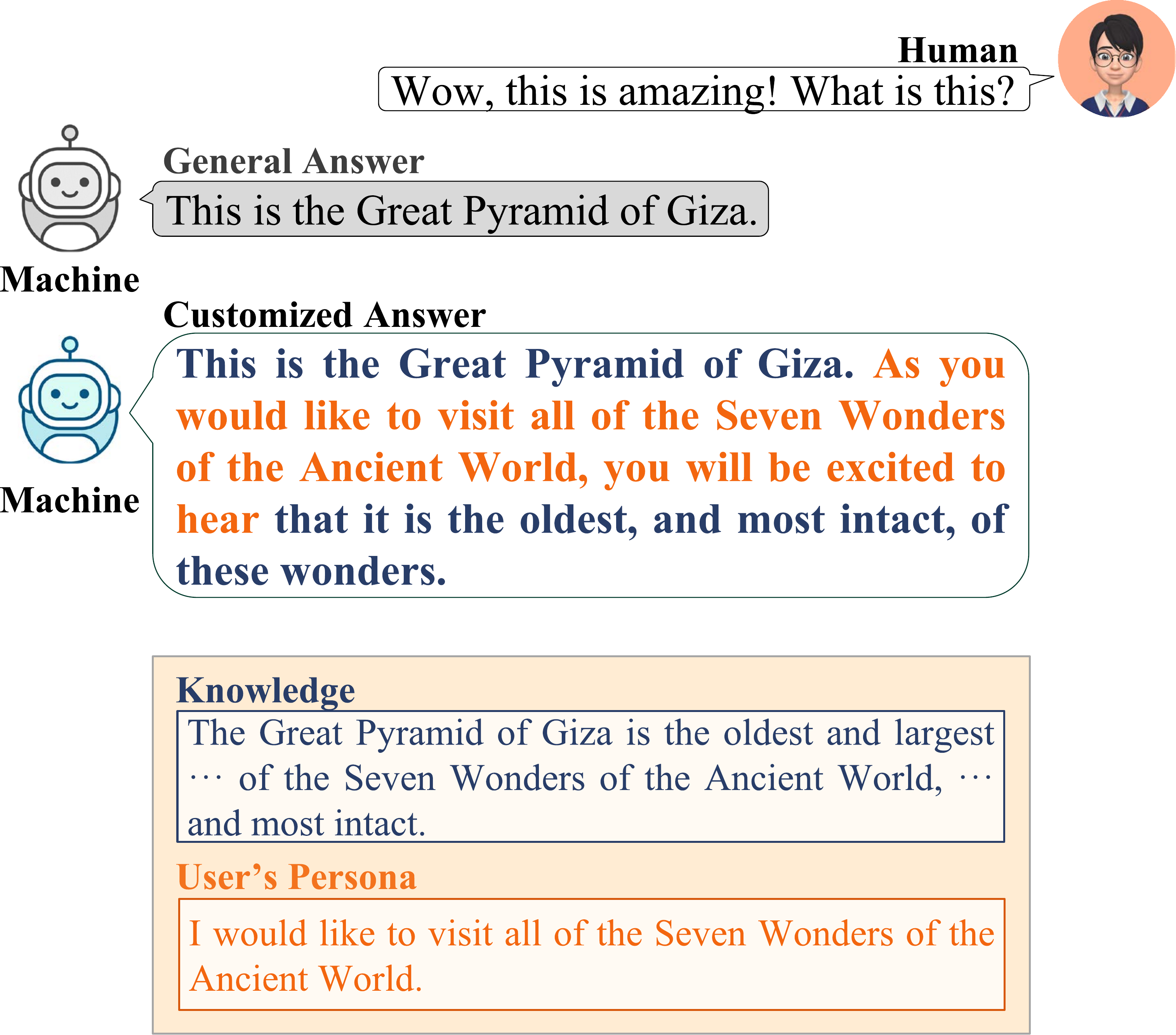}
	\caption{
	Objective of FoCus dataset.
	In contrast to the general answer, which only gives basic information, the machine's answer of FoCus dataset is more knowledgeable and customized, reflecting both knowledge and persona.}  
\label{fig:fig1}
\end{figure}

A person who is asked by a vegetarian to suggest a restaurant in New York City would not usually recommend Wolfgang's Steakhouse. When people give information to others, they consider the background of the person whom they are talking to. Following this manner of humans' conversation, a conversational agent's ability to have a conversation with \emph{customized answers} from prior knowledge and user's personal information is crucial for satisfying the users. For example, as exemplified in~\figref{fig:fig1}, the answer that considers both the user's persona and knowledge is much more attractive as well as informative.

Research for human-machine dialog has achieved significant success recently, owing to the advance of diverse dialog datasets~\cite{adiwardana2020towards, zhang2019dialogpt, shuster2019dialogue, li2017dailydialog, lowe2015ubuntu} and pre-trained language models~\cite{raffel2019exploring, clark2020electra, brown2020language}. Despite the remarkable success, the model's ability to give knowledge-grounded answers reflecting user's personal information remains largely limited. 

There exist several datasets and models that consider the user's persona, such as preference, interest or experience~\cite{majumder2020like, xu2020neural, wu2019guiding, zhang2018personalizing, rashkin2018towards, shuster2018image, li2017dailydialog, joshi2017personalization}, which contributes to building an agent that can talk about the user's feelings and interests. Though the dialog agent can access to the persona, the absence of knowledge often limits its ability of generating answers with specialized knowledge.

\begin{figure*}[ht]
	\centering
	\includegraphics[width=0.9\textwidth]{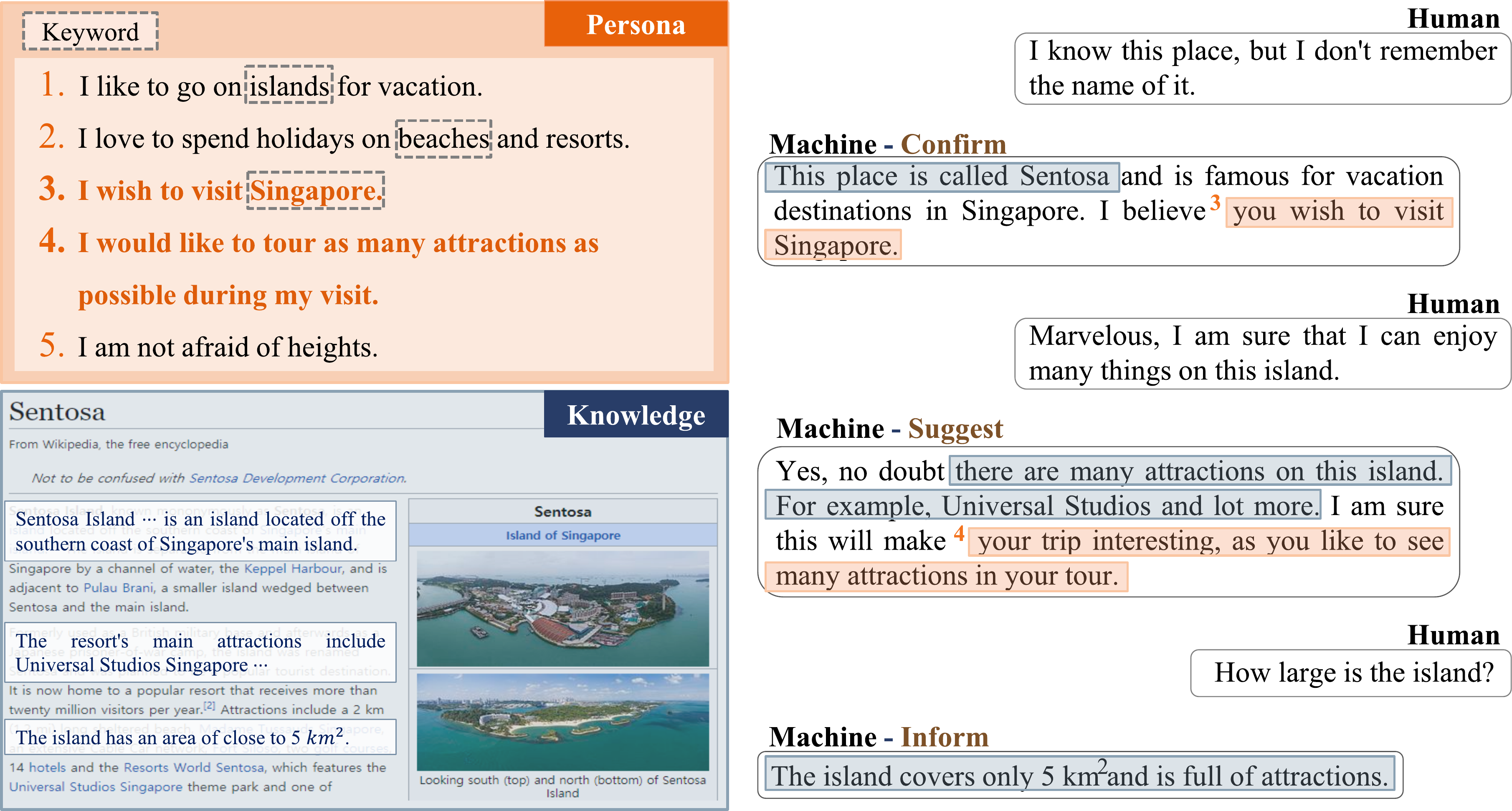}
	\caption{Example dialog between Human and Machine in FoCus dataset. The Human first asks about the landmark and the Machine then generates the answer considering the user's persona and Wikipedia knowledge. Answers can be made only with Wikipedia knowledge or both persona and Wikipedia knowledge. For instance, the third answer provides information about the size of the island only with knowledge. However, the second answer reflects both persona and knowledge.}  
\label{fig:fig2}
\end{figure*} 

Meanwhile, to build a dialog agent that generates more knowledgeable answers, datasets with the informative answers have been released~\cite{dinan2018wizard, zhou2018dataset}. In these datasets, the dialog agents learn to retrieve the required knowledge from the document. However, these datasets do not consider the user's persona, which restrict generating customized answers. Providing a large amount of knowledge without considering the user's background may result in giving the user useless information because people may need different types of knowledge, depending on their interests. 

For the ability to make use of both persona and knowledge, there have been a few attempts to blend them~\cite{smith2020can, roller2020recipes}. However, they merely stitch up the existing datasets, thus the models process only one source at a time, not both of them. Little work had been done on fusing the persona and knowledge into the utterances, thus there could not be sufficient conditions to build customized and intelligent conversational agents. 

In this work, we introduce a new dataset, \textit{call For Customized conversation} dataset\footnote{\url{http://github.com/pkchat-focus/FoCus}} (called FoCus), that supports knowledge-grounded answers that reflect user's persona. One of the situations in which people need different types of knowledge, based on their preferences, occurs when they travel around the world. As the knowledge of the landmark encompasses the range of history, design, structure, usage, tourism, and geological information, the diversity of the knowledge ensures. Inspired by this situation, we built a dataset where the agent informs the knowledge about the geographical landmark considering the user's persona. 

Our contributions are as follows:
\begin{itemize}
\item {We present the FoCus dataset in which the utterances contain both knowledgeable and customized answers for the first time.}

\item {We propose the baseline generative models trained on our dataset and evaluate them with the automatic scores and conduct human evaluation in respect to the generation abilities.}

\item {We provide two sub-tasks to measure the grounding ability, such as persona grounding (PG) and knowledge grounding (KG).}

\end{itemize} 

\begin{table*}[t]
\centering
\normalsize
\resizebox{\textwidth}{!}{
    \begin{tabular}{cccccc}
    \toprule
     & Knowledge Source & Persona Source
     & \# Dialogs & \# Average Turns 
     & \# Utterances \\
    \midrule

    Wizard of Wikipedia~\cite{dinan2018wizard} & \ding{51} &  \ding{55} & 22,311 & 9.0 & 201,999 \\
    CMU-DoG~\cite{zhou2018dataset} & \ding{51} & \ding{55} & 4,112 & 31.0 & 130,000 \\
    PERSONA-CHAT~\cite{zhang2018personalizing} & \ding{55} & \ding{51} & 10,907 & 14.0 & 164,356 \\
    \midrule
    \textbf{FoCus (Ours)} & \ding{51} & \ding{51} & 14,452 & 11.99 & 173,424 \\
    \bottomrule
    \end{tabular}
}
\caption{Comparison of our FoCus dataset with other datasets. Our dataset is composed of 14,452 dialogues, which has 12 average turns, with 173,424 utterances. The utterances of FoCus dataset consider both knowledge and persona sources.}
\label{tab:ExistingDataset}
\end{table*}

\section{FoCus Dataset}
To cover the diverse domain of a specific topic, we put the dialog under the setting of talking about Wikipedia knowledge on geographical landmarks. As the document of given landmarks provides various information of diverse domain, our dataset is well applicable to situations where the specialized knowledge is required. In this section, we describe the data collection process and analysis of the collected data. Also, we show three types of customized answers observed in our dataset. 

\subsection{Dataset Creation}
We collected the conversations about the geographical landmark guidance through Amazon Mechanical Turk (AMT)\footnote{We gave the qualification test to the workers for a high-quality dataset and paid 166 qualified workers \$5.5 for a single dialog.}. For the topic of dialogs, we selected a landmark from Google Landmarks Dataset v2 (GLDv2)~\cite{weyand2020google}. There are 5,316 Wikipedia pages on diverse landmarks, which have over 8,000 characters of contents to have abundant topics including history, design, tourism, and structures, and etc. For the persona sentences, we have 27,170 unique persona sentences related to landmarks' keywords implying its diversity.
We provided a corresponding Wikipedia page as a knowledge source to the workers. To select out the pages with abundant descriptions about diverse aspects of the topic We only adopted the pages of which the number of the characters is over 8,000. The workers were instructed with two-step data creation procedure: \textbf{Step 1. Make a Persona} and \textbf{Step 2. Make a Dialog}.

\paragraph{Step 1. Make a Persona.}
In the FoCus dataset, we define \texttt{persona}, described by five sentences, as a personal background which can be any sentence about experience, preference, possession, hobby or interest. The workers were instructed to choose their own avatar and landmark. Then they make a virtual personal background regarding the landmark. To encourage the workers to generate topic-relevant persona, we let them to extract the keywords in the given Wikipedia page and make the persona sentences by means of the keywords. By creating persona based on the keywords, the topic and persona become closely related, which leads to more engaging dialog, as exemplified in~\figref{fig:fig2}. Meanwhile, the workers were also allowed to create topic-agnostic persona sentences.

\paragraph{Step 2. Make a Dialog.}
After creating persona sentences, the workers were instructed to make a dialog by considering both persona and landmark knowledge. Unlike procedures done in previous datasets~\cite{dinan2018wizard,zhang2018personalizing}, they were instructed to make a multi-round dialog alone by alternating roles of \textit{human} and \textit{machine}, which enables more consistent and natural dialogs. We conducted the pilot study on the settings of creating dialog and concluded that the data from the single-person setup had high quality, especially in fusing persona and knowledge. As the person who asks the question knows better what knowledge one needs than the other person, the data from the single-person setup provided relevant and more customized answers.

To make customized and knowledgeable utterances, we gave the situation where the human asks a question regarding the landmark to the workers. In this situation, the machine answers the question by considering both \texttt{knowledge} and \texttt{persona} or only \texttt{knowledge}. As the human asks a question about the landmark which requires specialized knowledge to be answered, \texttt{persona}-only answer does not appear, which cannot give knowledgeable information to the user. For the first turn, we randomly gave one of the pre-generated questions so as to help the workers to smoothly start the first utterance of the dialog.

In addition, we also collected the grounding sources of machine's answers by letting the workers mark the sources they used, from \texttt{persona} or \texttt{knowledge}, when making answers. For instance, if they used \texttt{persona}, corresponding persona sentence was marked, and if they used Wikipedia \texttt{knowledge}, they indicate the referenced sentences in the Wikipedia page. These grounding information is used to evaluate the ability of models to ground the sources of their answers. The grounding abilities of the models can be quantitatively measured by proposed persona grounding (PG) and knowledge grounding (KG) sub-tasks, which will be described in the Experiments section.

\subsection{Dataset Analysis}
We report the comparison between our dataset and others with detailed statistics. In addition, characteristics of the customized answers in our dataset are analyzed.

\begin{table}[b]
\centering
\normalsize 
\resizebox{\columnwidth}{!}{
    \begin{tabular}{cccc}\\
    \toprule
     & Train & Valid & Test \\
    \toprule
    \# Dialogs & 11,562 & 1,445 & 1,445 \\
    \# Average Rounds & 6.00 & 6.00 & 5.99 \\
    \midrule
    Avg. Length of \textit{Human}'s Utt. & 40.94 & 40.89 & 41.08 \\
    Avg. Length of \textit{Machine}'s Utt. & 141.13 & 145.42 & 146.67 \\
    \midrule
    \# Knowledge-Only Answer & 35,580 & 4,501 & 4,437 \\
    \# Persona-Knowledge Answer & 33,792 & 4,169 & 4,225 \\
    \midrule
    \# Landmarks & 5,082 & 1,305 & 1,299 \\
    \bottomrule
    \end{tabular}}
\caption{Statistics of FoCus dataset.}
\label{tab:FoCusStat}
\end{table}

\subsubsection{Dataset Statistics}
We finally collected 14,452 dialogs with about 6 rounds per dialog on average. A comparison of our FoCus dataset with others is shown in~\tabref{tab:ExistingDataset}, including the number of dialogs, average turns, utterances, and data sources used. 
We split the collected data into train, valid and test sets. The average length of the machine's utterances, which is about 141.13 in the train set, is much longer than that of the human's, which is about 40.94. It is because the machine provides the specialized knowledge when answering the question. Also, 44,518 of knowledge-only answers and 42,186 of persona-knowledge answer. The detailed statistics of our dataset are summarized in~\tabref{tab:FoCusStat}.

\subsubsection{Types of Customized Answers}
The machine's answers can be categorized into three types according to their intent, i.e., \emph{Inform}, \emph{Confirm}, and \emph{Suggest}. We describe the characteristics of each intent type. Note that Utt. stands for Utterances. 

\paragraph{Inform.}
The answers that do not reflect the persona could be classified into \emph{Inform}, which is similar to types of previous dialog datasets~\cite{zhou2018dataset}. This type of answers only utilizes the knowledge when making an answer. As exemplified in~\figref{fig:fig2}, the answer that provides the size of the island is one of the examples. 

\paragraph{Confirm.}
The intent of \emph{Confirm} is to rephrase the user's persona and express the consent to it, as depicted in the first answer in~\figref{fig:fig2} . The answer of the machine confirms the user's preference for visiting Singapore. This type of answer is relatively more engaging than the answers with the \emph{Inform} intention, as the given persona sentences are reflected. They are similar to the answers from \citet{zhang2018personalizing}. However, these answers still have a limited range, and the persona is not deeply utilized in the answers.

\paragraph{Suggest.}
Unlike above two types of answers, the answers with \emph{Suggest} type recommends additional information that the users might like and enjoy or not suggest certain knowledge that users might hate or uncomfortable. This kind of answers give customized knowledge to the user by considering their persona, and they have not been introduced in other datasets. For example, the machine's answer that recommends the Universal Studios, because the user enjoys attractions during a tour, has the \emph{Suggest} intention.

\section{Model}
We introduce the baseline models trained on our FoCus dataset, consisting of a \emph{retrieval module} and a \emph{dialog module}. The \emph{retrieval module} retrieves the knowledge paragraphs related to a question, and the \emph{dialog module} generates utterances of the machine by taking the retrieved knowledge paragraphs, human's persona, and previous utterances as inputs. An overview of our model is depicted in Figure \ref{fig:fig3}.


\subsection{Notation}
The FoCus dataset is comprised of $N$ dialogs and each dialog is composed of $R$ rounds such that $D=\{(u_{1}^{h}, u_{1}^{m}), ..., (u_{R}^{h}, u_{R}^{m})\}$ with the utterances of human $u^{h}$ and machine $u^{m}$. The dialog is given with the corresponding persona and landmark knowledge. The human's persona is denoted as $P$, and knowledge documents about landmark are indicated as $K$. We further define the candidate sets of persona and knowledge, $C_{P}$ and $C_{K}$, respectively, which are given at every turn and composed of the ground truth answers and distracting answers. Such candidates can be used to improve the grounding ability of agent by learning to select a ground truth answer among them, and more details are in Experiments section. The number of candidates of $C_{P}$ and $C_{K}$ are $J$ and $S$, respectively.

\subsection{Retrieval Module} 
\begin{figure*}[t]
	\centering
	\includegraphics[width=0.9\linewidth]{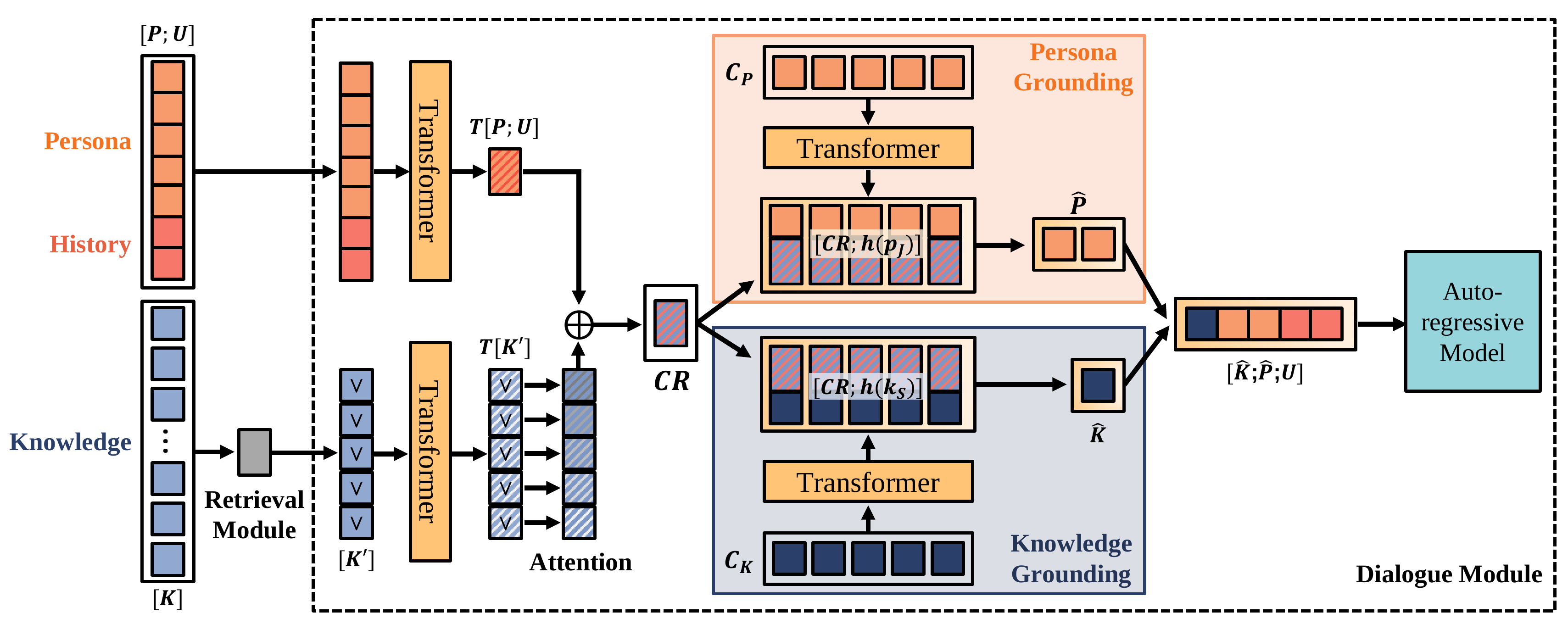}
	\caption{
	Overview of model architecture. The retrieval module selects five paragraphs $K'$ from the documents of the given landmark. It goes through Transformers and is updated with attention mechanism. It is concatenated with the representation of Transformer-encoded sequence of persona and history, depicted as a cross in a circle. The CR is trained for the grounding tasks, and chosen persona and knowledge ($\hat{P}$ and $\hat{K}$) from the given candidate sets ($C_{P}$ and $C_{K}$) are used to train the model's generation competence.}  
\label{fig:fig3}
\end{figure*}

To avoid excessive memory consumption, we present a retrieval module that enables narrowing the Wikipedia document down to five paragraphs $K'$, which are related to the given utterance of human $u^{h}$. Among KNN \cite{fix1989discriminatory}, TF-IDF \cite{salton1988term} and dense passage retrieval methods, we choose the TF-IDF score to retrieve the most related top 5 passages for the fast and efficient computation. To ensure its retrieval capability, BERTscore~\cite{zhang2019bertscore} is used to estimate how much the retrieved paragraphs are semantically similar to the gold knowledge. Note that the gold knowledge is reconstructed by the workers with their chosen sentences from the given Wikipedia paragraphs. The average BERTscore between the gold knowledge and the top 1 paragraph is about 83\%, which is a relatively high score. As a result, TF-IDF score is used to choose five paragraphs, $K'$, from the given knowledge document $K$ which is utilized as the knowledge source for the answer, as shown in \figref{fig:fig3}. We calculate term frequency-inverse document frequency (TF-IDF) similarity score between the last question of human and possible knowledge paragraphs after the evaluation on the retrieved paragraph. The average token number of retrieved passages is about 132, and only the first 150 tokens are used as inputs.


\subsection{Dialog Module}
After selecting relevant knowledge paragraphs, the model first generate context-relevant representations to obtain the vectors that is highly relevant to the given knowledge, persona, and history. The representations are used to select the persona and knowledge from $C_{P}$ and $C_{K}$, respectively. Chosen knowledge and personas are then concatenated with the dialogue history and then fed into the language modeling along with the machine's answer. Consequently, our training objectives are composed of language modeling for persona grounding, knowledge grounding and utterance generation among the given persona and knowledge candidates, $C_{P}$ and $C_{K}$, which is trained in a multi-task learning (MTL) fashion~\cite{ruder2017overview, zhang2021survey}. The number of candidates, $J$ and $S$, are 5 and 10 respectively.


\paragraph{Context-Relevant Representation.}
Dialog module first makes a Context Relevant representation ($CR$) of the current dialog turn. Chosen knowledge paragraphs $K'$ and a concatenation of persona and history $[P;U]$ are given as inputs. They are encoded by transformer, resulting $T(K')$ and $T([P;U])$ respectively, where $T$ denotes a transformer model. Then, $T(K')$ is updated with the attention~\cite{bahdanau2014neural} mechanisms and concatenated with $T([P;U])$ resulting in the final representation $CR$.

\paragraph{Persona Grounding.}
To make a model that reflects the proper persona of the human when making answers, the model learns which persona to utilize, given the $CR$ representation. As multiple persona sentences or none of them could be in the ground-truth answers, we train our model to discriminate each persona sentence to be used among the persona candidates. The special tokens are added to the each candidates. We utilize them by concatenating $CR$ and the last hidden state representations of the special tokens from each candidate. The loss function is defined as follows:
\begin{equation}
\begin{split}
    {L}_{PG} = &-\sum_{j=1}^{J} (q^*_j \textrm{log} \ \mathrm{Prob}([CR;h(p_j)]) \\ 
    &+  (1 - q^*_j) \textrm{log} \ (1 - \mathrm{Prob}([CR;h(p_j]))),
\end{split}
\end{equation}
with $q^*_j$ denoting a label defined as 1 if $j$-th persona sentence is ground-truth, 0 otherwise. $h(p_{j})$ is the last hidden state representation of the special token of $p_{j}$. $\mathrm{Prob}([CR;h(p_j)])$ is the estimated probability of the models. 

\paragraph{Knowledge Grounding.}
The model also learns to use knowledge grounding to generate informative answers. The $C_{K}$ consists of the ground-truth sentence and distracting candidates that are from the documents of different landmark. Given knowledge candidates at each round, the model is trained to choose one knowledge item that is expected to be used to answer the question by concatenating $CR$ and the last hidden state representations of the special tokens from knowledge candidates: 
\begin{equation}
    {L}_{KG} = -\sum_{s=1}^{S} q^*_s \textrm{log} \
    \mathrm{Prob}([CR;h(k_{s})]),
\end{equation}
with $q^*_s$ denoting a label defined as 1 if $s$-th knowledge paragraph is ground-truth, 0 otherwise. $h(k_{s})$ is the last hidden state representation of the special token of $k_{s}$. $\mathrm{Prob}([CR;h(k_{s})])$ is the estimated probability of the models. 

\paragraph{Language Modeling.}
To build a generative agent, we model the machine's utterances in an auto-regressive manner. We consider two types of model structures, that are decoder-only and encoder-decoder. Following the previous works of \citet{jelinek1980interpolated, bengio2003neural}, the language modeling loss function is defined such that 
\begin{equation}\label{encoder-decoder}
    {L}_{LM} = -\sum_{i=1}^{I} \textrm{log} \ \mathrm{Prob}(x_{i}|v, x_{1}, ..., x_{i-1}),
\end{equation}
where $\mathrm{Prob}(\cdot)$ denotes a probability of the langauge model, $x_{i}$ is $i$-th token of $u^{m}$, $I$ is the number of tokens and $v$ stands for the sequence $[\hat{K};\hat{P};U]$ with concatenation of $\hat{K}$, $\hat{P}$, and $U$. $\hat{K}$ and $\hat{P}$ are the predicted candidates by the model in the knowledge grounding (KG) and persona grounding (PG) tasks respectively. Note that in the decoder-only model, $[\hat{K};\hat{P};U]$ are defined as the sequence of previous tokens, while they are used as the encoder inputs in the encoder-decoder model.

\paragraph{Full Objectives.}
The entire loss function aims to minimize the negative log-likelihood of language modeling and sub-tasks as in \cite{radford2019language, wolf2019transfertransfo}. The full training objectives are defined as follows:
\begin{equation}
\label{full-training}
    {L} = \lambda_{PG} {L}_{PG} + \lambda_{KG} {L}_{KG} + \lambda_{LM} {L}_{LM},
\end{equation}
where $\lambda$ controls the proportion of each task during the training. In the experiments, $\lambda_{LM}$, $\lambda_{PG}$, and $\lambda_{KG}$ were set to 10, 1 and 1, respectively. $\lambda$ is chosen by the manual search.

\begin{center}
\begin{table*}[ht]
\normalsize 
\centering
\renewcommand{\arraystretch}{1.0}
\begin{tabular}{l|cccccc|cc}
\toprule
\multicolumn{1}{c|}{\multirow{2}{*}{Models}} & \multicolumn{6}{c|}{Generation} & \multicolumn{2}{c}{Grounding (Acc.) } \\ \cline{2-9} 
\multicolumn{1}{c|}{} & \multicolumn{1}{c}{PPL} & \multicolumn{1}{c}{chrF++} & \multicolumn{1}{c}{BLEU} & \multicolumn{1}{c}{R-1} & \multicolumn{1}{c}{R-2} & \multicolumn{1}{c|}{R-L} & Persona & Knowledge \\
\hline
\hline
Decoder +PG +KG & 228.69  & 0.1565 & 3.53 & 22.41 & 4.78 & 18.60 & \textbf{67.83} & 64.28 \\
\hline
Enc-Dec +PG +KG & 428.75 & 0.1345 & 2.79 & 18.45 & 2.81 & 14.80 & \textbf{67.83} & 64.52 \\
\hline
GPT-2 & 17.42 & 0.1942 & 5.97 & 26.61 & 9.73 & 23.13 & 65.50 & 10.71 \\
GPT-2 +PG  & 18.45 & 0.2221 & 5.63 & 25.56 & 9.12 & 22.20 & \textbf{67.83} & 9.25 \\
GPT-2 +KG  & \textbf{10.73} & 0.2875 & 11.29 & 36.35 & 19.89 & 32.35 & 45.61 & 71.33 \\
GPT-2 +PG +KG & 11.45 & 0.2777 & 10.65 & 35.26 & 18.82 & 31.33 & \textbf{67.83} & 70.95 \\
\hline
BART & 26.55 & 0.1982 & 5.70 & 25.67 & 8.90 & 21.70 & 67.49 & 14.05 \\
BART +PG & 26.54 & 0.1932 & 5.36 & 25.35 & 8.43 & 21.40 & \textbf{67.83} & 14.75 \\
BART +KG & 15.84 & \textbf{0.2946} & \textbf{11.64} & \textbf{36.19} & \textbf{19.90} & \textbf{31.84} & 53.78 & \textbf{73.00} \\
BART +PG +KG & 23.25 & 0.2887 & 11.28 & 35.35 & 19.12 & 31.06 & \textbf{67.83} & 71.70 \\
\bottomrule
\end{tabular}

\caption{Experimental results of the baseline models on the test set. The models are evaluated by generation metrics, including perplexity (PPL), chrF++, SacreBLEU, ROUGE-1 (R-1), ROUGE-2 (R-2) and ROUGE-L (R-L), and accuracy for persona grounding task and knowledge grounding task.}
\label{tab:experiment}
\end{table*}
\end{center}

\section{Experiments} 
In this section we describe all the details of experiments including baselines, training settings and evaluation. We also analyze the experimental results and human evaluation of the dialog models trained on our dataset.

\subsection{Language Model Baselines}
We first describe the baseline language models, including transformer decoder, transformer encoder-decoder, GPT-2 and BART. By being trained with multi-tasks, those models are able to choose which persona and knowledge to use, as well as generate utterances. We implement the models based on the source code of HuggingFace's transformers~\cite{wolf2020transformers, wolf2019transfertransfo}.

\paragraph{Transformer-based Models.}
We train the models with a transformer~\cite{vaswani2017attention} structure. Both decoder-only model and encoder-decoder model are used to generate the utterances. To evaluate the effectiveness of pre-training, we set transformer layers to have the same structure with the following pre-trained language models.

\paragraph{Pre-trained Language Models.}
We adopt GPT-2~\cite{radford2019language} and BART~\cite{lewis2019bart} as pre-trained decoder-only and pre-trained encoder-decoder models, respectively, which are known to show remarkable performances in language generation by training a colossal number of parameters on a massive corpus.

\subsection{Experimental Setup}
We train GPT-2$_{Small}$, which has 12 layers and 12 attention heads with 768 embedding dimensions, and BART$_{Base}$, which has 6 layers each in both the encoder and decoder, and 12 attention heads with 768 embedding dimensions. We use a batch size of 4 with a gradient accumulation of 32. Adam optimizer is used, and the learning rate is set as 6.25e-5, where $\beta_{1}$ = 0.9, $\beta_{2}$ = 0.999 with linear decay. For the hyperparameter settings, we adopt the initial hyperparameters from the models trained on PERSONA-CHAT~\cite{zhang2018personalizing} and Wizard-of-Wikipedia~\cite{dinan2018wizard} datasets. Among the candidates, we choose the hyperparameters that showed the best performance. Fine-tuning them on the entire data with 2 epochs takes approximately 10 hours with one RTX-8000 GPU. For the utterance generation, we use the nucleus sampling with top-p = 0.9 and sampling temperature with 0.7. The maximum sequence length is set to 20. Generation and grounding evaluation takes about 30 minutes. 

\begin{center}
\begin{table}[b]
\centering
\normalsize 
\resizebox{\columnwidth}{!}{
    \renewcommand{\arraystretch}{1.0}
    \begin{tabular}{lcccc}
    \toprule
    Model & Rank & Fluency & Engagement & Consistency \\
    \toprule
    Human & \textbf{1.05} (0.31) & \textbf{4.15} (1.54) & \textbf{4.08} (1.53) & \textbf{4.06} (1.47)  \\
    \midrule
    GPT-2 & 2.64 (0.48) & 2.85 (0.93) & 2.95 (0.98) & 2.76 (0.99) \\ 
    \midrule
    BART & 2.31 (0.52) & 3.13 (1.14) & 3.18 (1.08) & 3.10 (1.04) \\ 
    \bottomrule
    \end{tabular}}
\caption{Human evaluation. The models trained with PG and KG are evaluated their utterances compared to the gold data of human. The value in the parenthesis indicates standard deviation of the scores.}
\label{tab:human_eval}
\end{table}
\end{center}

\subsection{Automatic Score Evaluation}
To evaluate model's ability to give fluent, attractive and informative utterances, sub-tasks for measuring the ability of generating customized responses (\emph{generation}) and discriminating which source to reflect (\emph{grounding}) are provided.  

\paragraph{Task 1 - Generation.} To evaluate the generation competence, the perplexity (PPL) is used to measure the fluency as in other generation tasks \cite{zhang2018personalizing, dinan2018wizard}. The chrF++ \citep{popovic2017chrf++} score, SacreBLEU \citep{post2018call}, and recall-oriented understudy for gisting evaluation (ROUGE-1-F, ROUGE-2-F, ROUGE-L-F) \cite{lin2004rouge} are adopted to assess how close the generated answer is to the original answer.

\paragraph{Task 2 - Grounding.} In addition, we evaluate the models' grounding abilities by our proposed PG and KG tasks, which enable us to test whether the models choose the proper persona and knowledge among the given candidates to generate an answer. As an answer of the machine that utilizes different persona and knowledge at each turn, we provide the persona candidates and knowledge candidates for every round. Whereas $C_{P}$ consists of five given persona sentences, $C_{K}$ includes the ground-truth sentences of Wikipedia and distracting candidates that have the same number of sentences from the other documents on different landmarks. We measure the accuracy of persona grounding and knowledge grounding persona selection and knowledge selection respectively.

\paragraph{Analysis.} As shown in \tabref{tab:experiment}, we experiment with transformer-based decoder model, encoder-decoder model, GPT-2 and BART. We analyze their generation abilities on the test set. Out of the transformer-based models, the decoder-only model shows higher generation performance than the encoder-decoder model. In the grounding task, they show comparable performances. GPT-2$_{Small}$ and BART$_{Base}$ models are adopted as pre-trained language models, and they are trained to generate the machine's utterances. To investigate the effectiveness of the grounding task, we additionally train the models with or without two grounding sub-tasks. In the generation task, the language models trained with knowledge grounding (KG) task show high scores, especially BART trained with KG is the highest on the most of generation scores. However, their persona grounding (PG) accuracy is lower than others, which means that they are not good at choosing proper persona for each turn. The language models trained with both PG and KG show slightly lower but comparable performances in the generation task and, but they show competent scores in both of two grounding sub-tasks. Since all the results are rounded to two decimal point, numbers from PG seem to be the same. The best results of PG are converged to a certain number and it indicates the upper bound of the baseline models. Also, our experimental results indicate that the high automatic score on the generation task does not always guarantee the high grounding ability. The experiments suggest the need of versatile generative models that are able to not only make fluent utterances, but also select proper sources and fuse them competently.

\begin{table}[t]
\centering
\normalsize
\resizebox{\columnwidth}{!}{
\begin{tabular}{|p{\columnwidth}|}
\hline
\textbf{Persona} \\
\hline
1. I live in a building \\
2. I find heritage-listed buildings interesting \\
3. I am from Australia \\
4. I have never been to Queensland \\
5. I wish to visit Queensland \\

\hline
\textbf{Landmark} \\
\hline
https://en.wikipedia.org/wiki/Thorps\_Building \\
\hline
\textbf{Dialog} \\
\hline
\textit{Human} : Cool! What is it? \\
\textit{(1) BART} : The is a beautiful urban burial ground that contains a collection of highly intact funerary monuments and furniture dating \\
\textit{(2) BART +KG} : This is Thorps Building, a heritage-listed commercial building. \\
\textit{(3) BART +PG} : It is a historic burial ground located in Australia where you are from. \\
\textit{(4) BART +PG +KG} : It is a heritage-listed commercial building, you may have heard of it since you are a fan \\
\hline
\end{tabular}}

\caption{An example of conversations between human and BART$_{Base}$. (1) generates fluent utterance, but it is not closely related to the given persona and knowledge. (2) makes an informative answers, and (3) generates more user-aware answer. (4) seems to generate the most plausible utterance by fusing both persona and knowledge.}
\label{table:example_conversation}
\end{table}

\subsection{Human Evaluation}
To evaluate the fluency, engagement, and consistency in the utterances of machine on a numerical scale from 1 to 5, we randomly selected 20 dialogues generated by the models which are in the test set. We set up three questions and specified the level of answers with likert scale~\cite{likert1932technique}. In addition, we asked human evaluators, the MTurk workers \footnote{We paid 4 qualified workers \$2 for a single evaluation on the dialog.}, to rank the each examples in order of which conversation shows the most human-like utterances by the machine following \citet{cho2020grounding}. Rank is scaled from 1 to 3 and the lower number indicates the better quality. The survey results are shown in \tabref{tab:human_eval}. The gold data made by human shows the best scores on all criteria of fluency, engagement and consistency, which ranks first. Among GPT-2 and BART, note that they are trained on PK and KG, BART is shown to outperform GPT-2 on the all criteria. The result shows that the quality of the gold data surpasses the models' generation. In spite of the pre-trained models' massive parameters and their abilities, their responses, given the context, are much less engaging, fluent and consistent than those of humans which means that our dataset is considerably challenging.

\subsection{Grounding Quality Assessment}
With the human evaluators, we evaluate the grounding quality of the dataset. We asked the workers to assess whether the answers in each utterance included Wikipedia knowledge or both Wikipedia knowledge and persona sentences. We had each dialogs evaluated by five independent workers \footnote{We paid 20 qualified MTurk workers \$2 for a single evaluation on the dialog.} with the randomly selected 200 dialogs in our dataset. The results in \tabref{tab:data_quality} shows the proportions of well grounded utterances. The proportions of well-grounded utterances with knowledge-only are about 99\% and and those of knowledge-persona grounded answers are over 94\%. 

\begin{table}[t]
\centering
\normalsize
\begin{tabular}{lc}
\toprule
Answer Type & Well-grounded utterances (\%) \\
\toprule
Knowledge-only & 98.94 \\ 
\midrule
Knowledge-Persona & 94.52 \\ 
\bottomrule
\end{tabular}
\caption{Grounding quality assessment. The numbers indicate the proportions of the well-grounded utterances with knowledge, and both knowledge and persona respectively.}
\label{tab:data_quality}
\end{table}

\section{Related Work}
To build dialog agents that can interact with people in multi-turn conversations, several datasets have been introduced~\cite{ ritter2010unsupervised, danescu2011chameleons, lowe2015ubuntu, wu2016sequential, li2017dailydialog, mostafazadeh2017image, shuster2018image, fan2019eli5}. Despite these datasets, the dialog agents merely answer the question without considering the user or specialized knowledge.

To generate customized answers to the users, attempts have been made to endow the agent with the user's emotion, preference, and experience ~\cite{rashkin2018towards, shuster2018image, urbanek2019learning, boyd2020large}. \citet{zhang2018personalizing} introduces a dataset that includes each speaker's preference and experience, where persona sentences describing two speakers are given. Because both speakers are only provided with persona sentences, one speaker simply confirms what the other speaker likes or dislikes in the dialog. Even though agents generate answers that react or express sympathy, they hardly give a document-grounded answer that fits the user's preference and experience. 

While the user-centered dialog datasets have appeared, datasets and agents that aim to improve the level of knowledge in the answer with additional documents has been in parallel released~\cite{dinan2018wizard, zhou2018dataset, moghe2018towards, qin2019conversing, gopalakrishnan2019topical, cho2020grounding, zhou2020kdconv, santhanam2020local}. \citet{dinan2018wizard} is a dialog dataset where the agent retrieves the Wikipedia pages on diverse topics and generates responses to the questions. Although these data have a concept of persona, they do not contain customized answers to the listener. Similar to \citet{dinan2018wizard}, \citet{zhou2018dataset} introduces a document-grounded dataset that includes specified documents from Wikipedia articles about popular movies. These datasets mainly consist of answering the question without considering the user's information, and it leads to excessive and needless answers. 

There have been efforts to blend several datasets~\cite{shuster2019dialogue, smith2020can} to build an intelligent agent which has multiple abilities learned from various datasets. Despite the previous datasets, the capability of machines to respond in a dialog is still insufficient, compared to that of humans. Specifically, to answer a question, retrieving the knowledge while considering the user's background information is beyond current dialog agent's abilities.

\section{Conclusion}
In this work, we have introduced the FoCus dataset that contains the customized responses by utilizing both persona and the Wikipedia knowledge. To validate the effectiveness of our dataset, we adopted and trained the language models on the FoCus dataset. Along with the generation tasks, we evaluate the grounding abilities of the models with provided PG, and KG sub-tasks. The experiments demonstrated that the pre-trained models show high performance on the generation task, but it does not necessarily lead to high grounding performance, and may limit in the grounding abilities. As shown in human evaluation and grounding quality assessment, our dataset is proven to be natural but complicated for the machines to mimic. We believe our FoCus dataset can contribute to build more human-like agents which gives customized answers with proper knowledge. We will also additionally annotate the type of the intents for each answer to let the models learn the purpose during generating answers. In the future, the models trained with our dataset can be utilized in the situation where the specialized knowledge is required depending on the user's persona in the form of personal assistants. We hope that the researches aim to make dialog agents more attractive and knowledgeable with grounding abilities to be explored.

\section{Acknowledgments}
This work was supported by Institute of Information \& communications Technology Planning \& Evaluation(IITP) grant funded by the Korea government(MSIT) (No. 2020-0-00368, A Neural-Symbolic Model for Knowledge Acquisition and Inference Techniques). This research was supported by Basic Science Research Program through the National Research Foundation of Korea(NRF) funded by the Ministry of Education(NRF-2021R1A6A1A03045425). Also, this work was supported by NCSOFT NLP Center. 

\bibliography{anthology,aaai22}

\clearpage
\appendix
\setcounter{secnumdepth}{1}

\section{Appendix}
\subsection{Pre-generated Initial Questions}
\subsubsection{}
\begin{table}[h]
\centering
\begin{tabular}{@{}c|l@{}}
\toprule
  & \multicolumn{1}{c}{Pre-generated Initial Question} \\ \midrule
1 & "Where is this place?" \\ \midrule
2 & "Wow, this is amazing! What is this?" \\ \midrule
3 & \begin{tabular}[c]{@{}l@{}}"I think I've been there before but \\ I don't remember the name of this place."\end{tabular} \\ \midrule
4 & \begin{tabular}[c]{@{}l@{}}"I know this place, but \\ I don't remember the name of this place."\end{tabular} \\ \bottomrule
\end{tabular}
\end{table}

\subsection{Data Collection Pages}
\begin{figure}[h]
	\centering
	\includegraphics[width=0.85\linewidth]{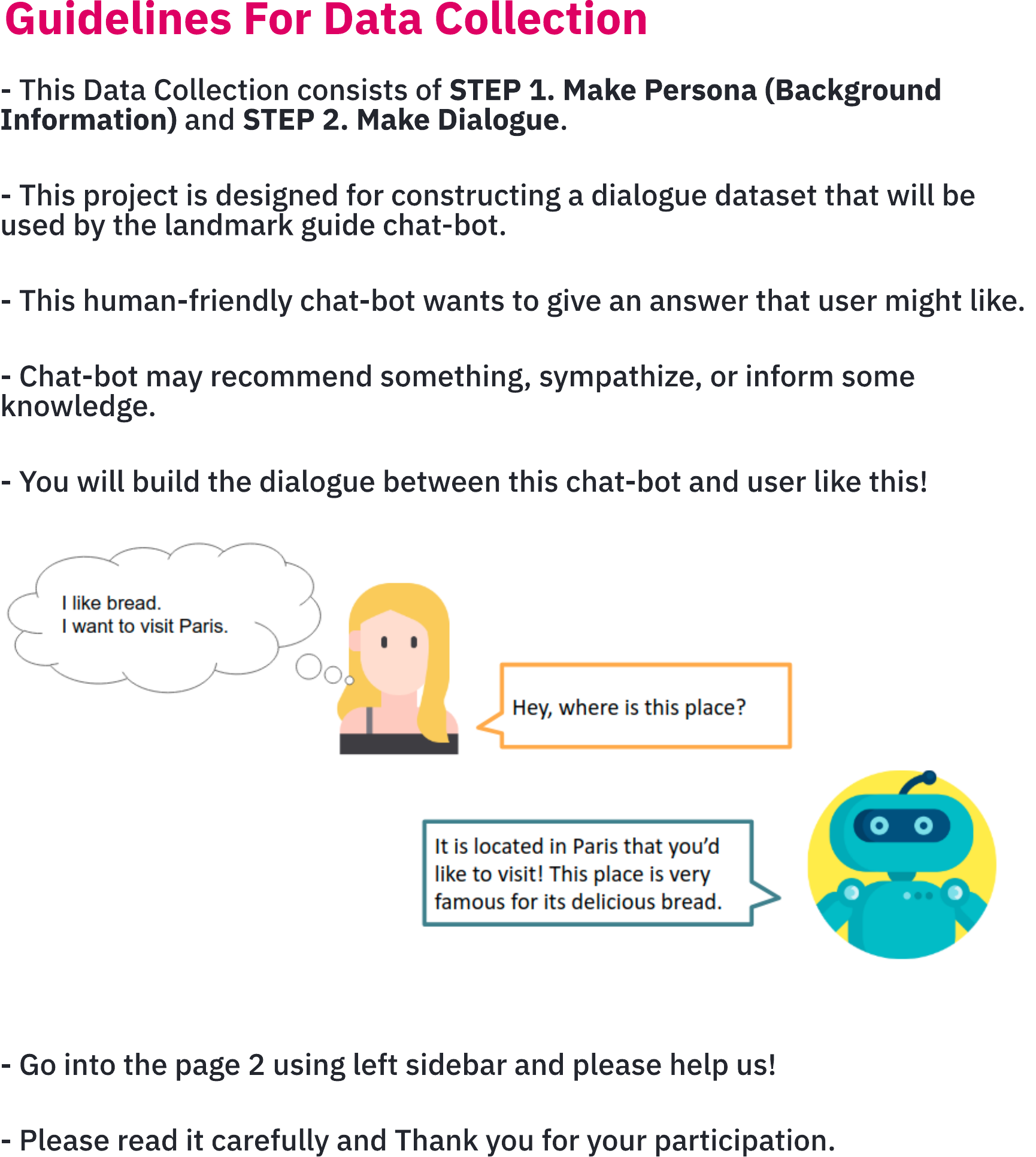}\\
\caption{Introduction to the data collection.}
\end{figure}

\begin{figure}[h]
	\centering
	\includegraphics[width=0.85\linewidth]{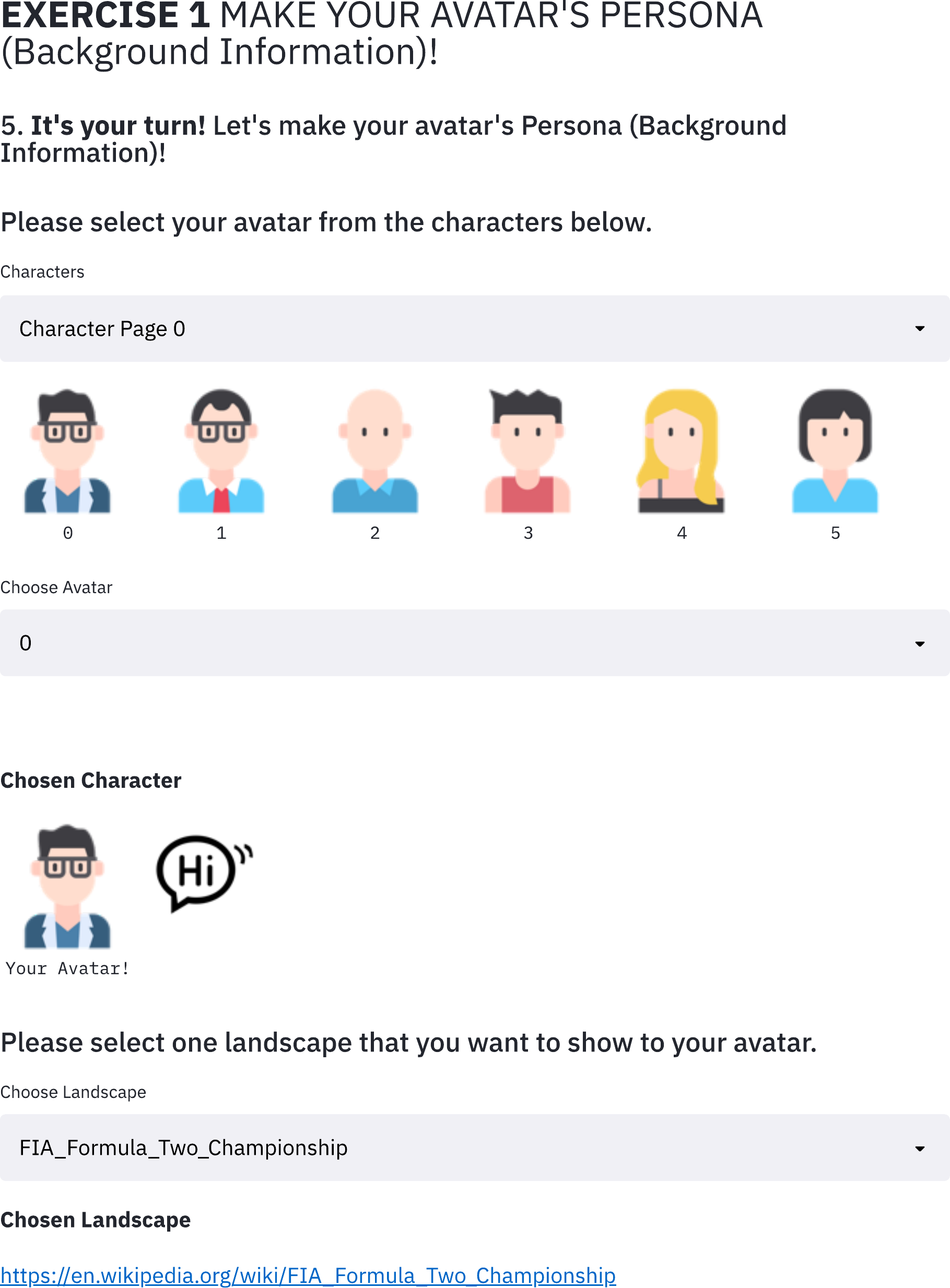}
\caption{Avatar and Landmark selection. The workers choose their avatar and landmark to make a dialog.}
\end{figure}

\begin{figure}[h]
	\centering
	\includegraphics[width=0.85\linewidth]{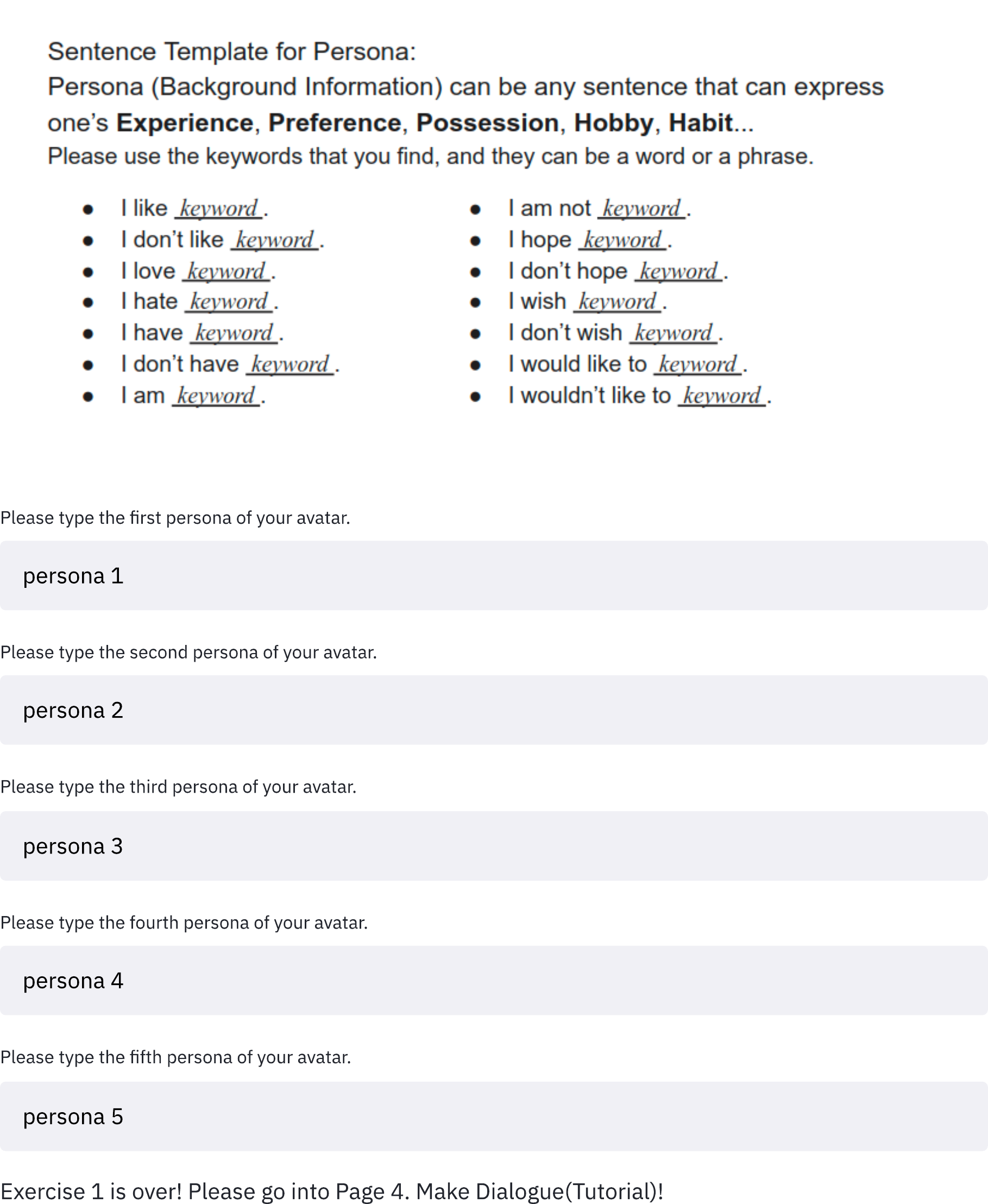}
\caption{Persona creation page. The workers make the five sentences of persona with the keywords from the Wikipedia page link.}
\end{figure}

\begin{figure}[h]
	\centering
	\includegraphics[width=0.85\linewidth]{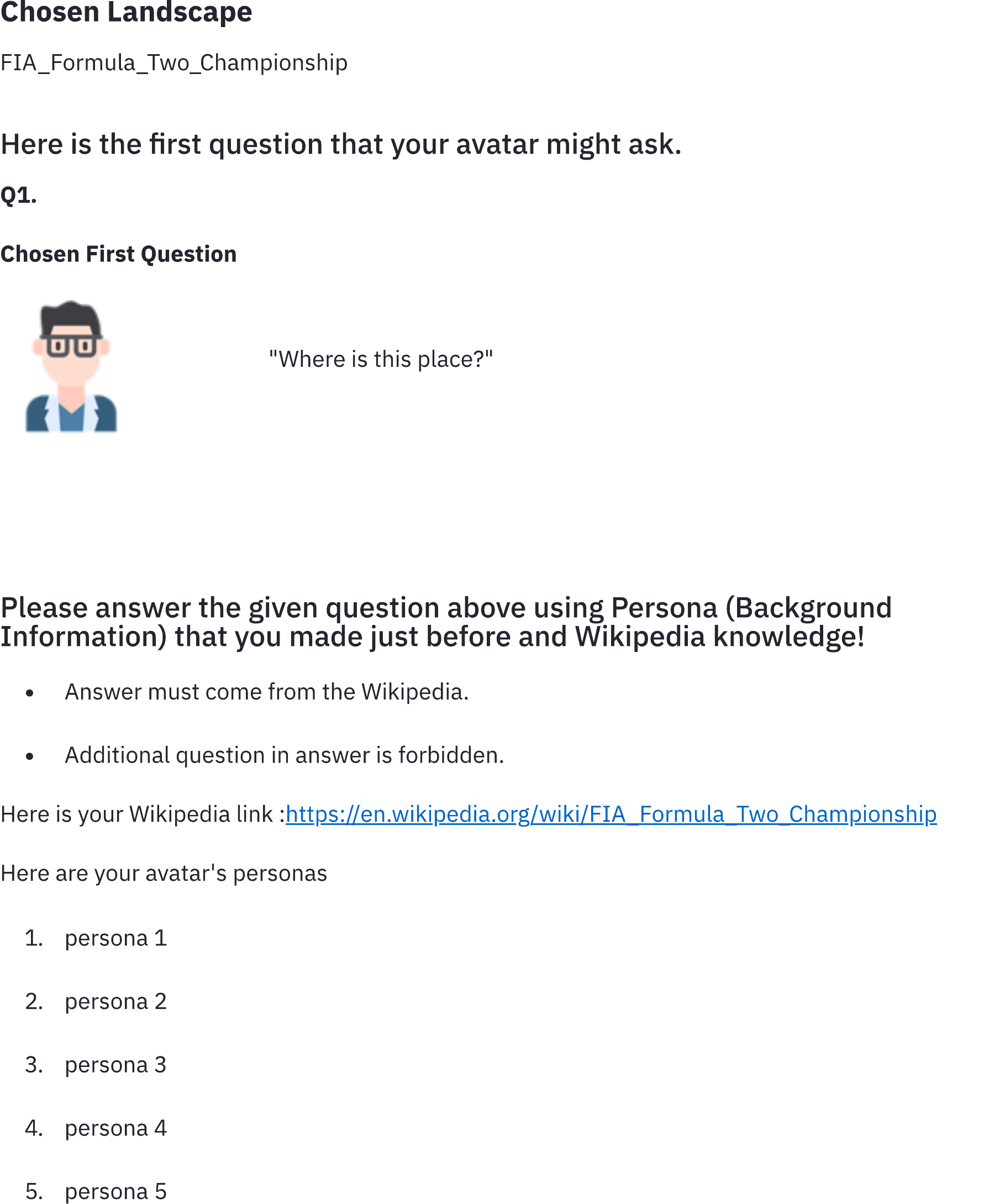}
\caption{First question from pre-generated question. The workers are given the first question}
\end{figure}

\begin{figure}[h]
	\centering
	\includegraphics[width=0.85\linewidth]{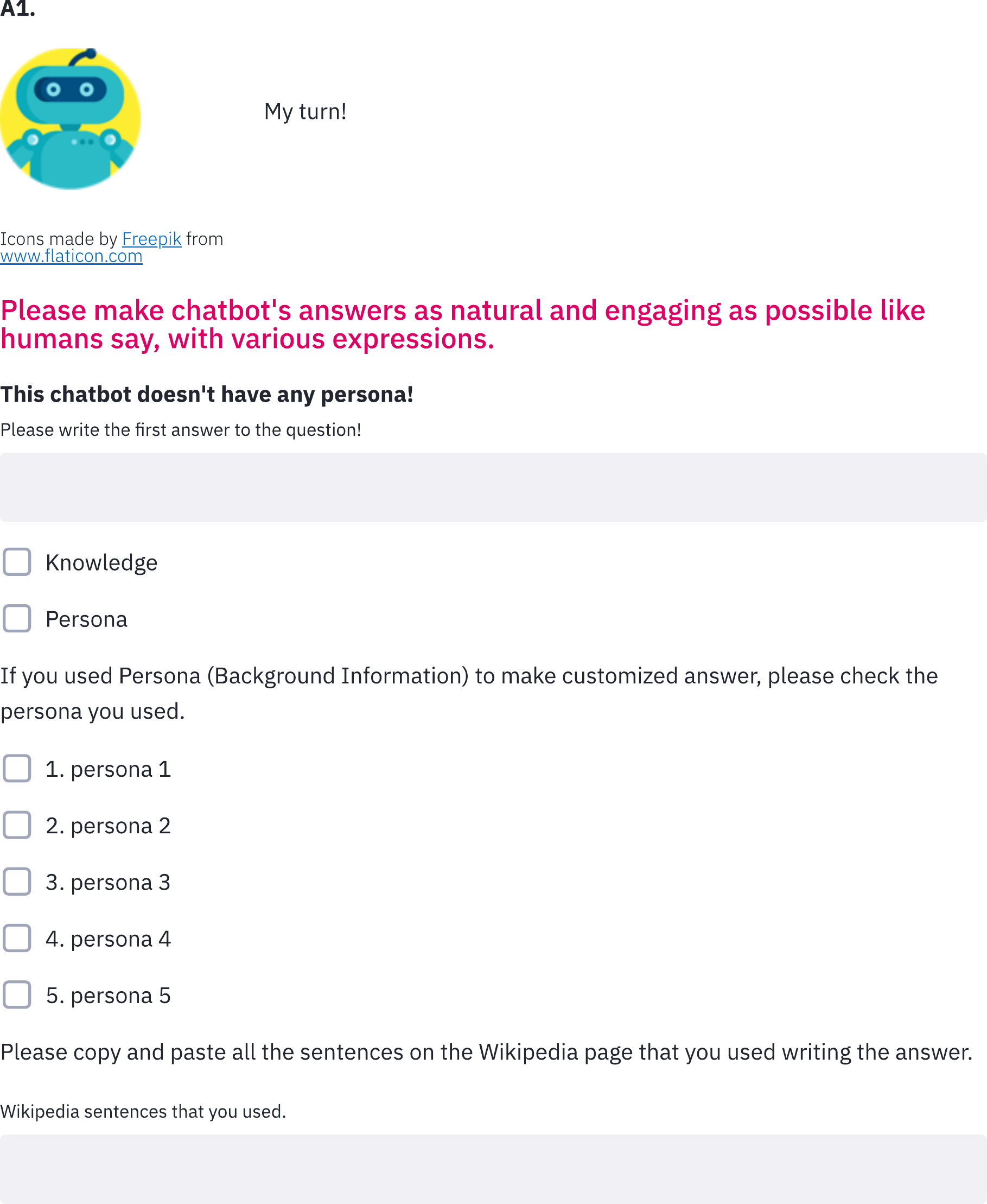}
\caption{First answer to the question. The workers make the first answer given the first question.}
\end{figure}

\begin{figure}[h]
	\centering
	\includegraphics[width=0.85\linewidth]{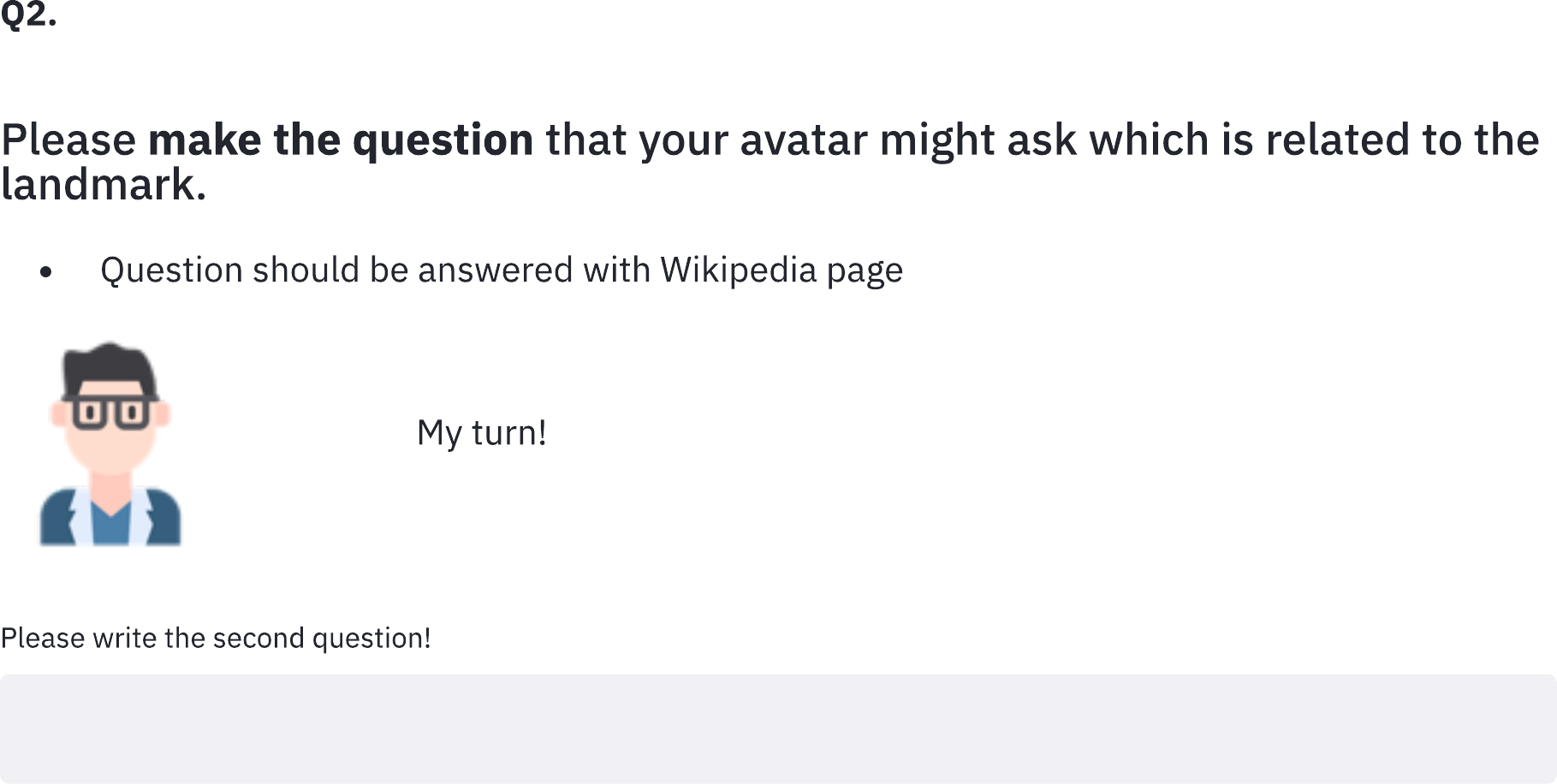}
\caption{Second question creation. Different from the first question, the workers make the second question to the last question by themselves. The question requires specialized knowledge about the landmark to be answered.}
\end{figure}

\begin{figure}[h]
	\centering
	\includegraphics[width=0.85\linewidth]{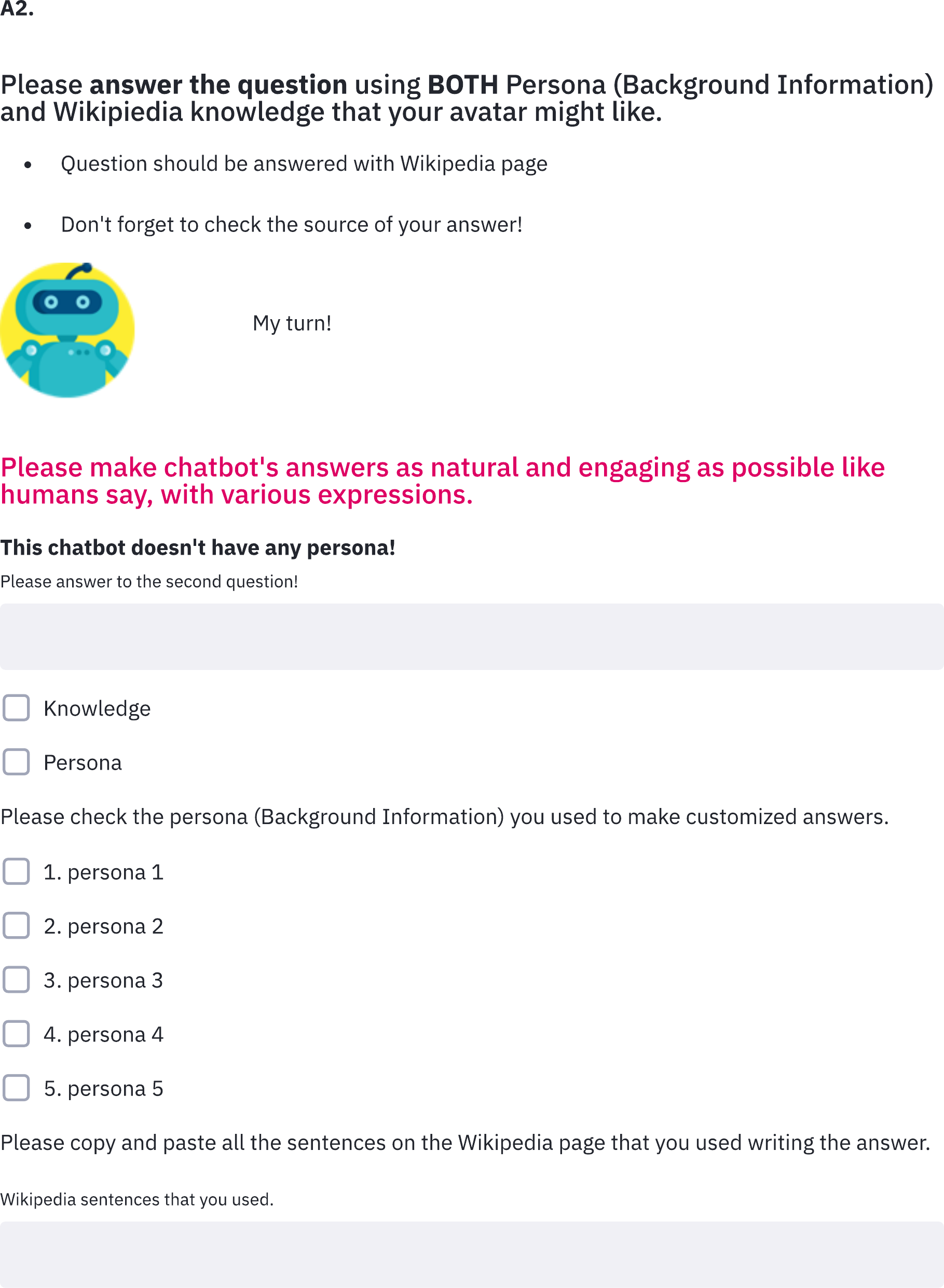}
\caption{Second answer to the question. From the second answer to the last answer, they have similar form with this page. The workers answer the question with the knowledge in the Wikipedia document and the user's persona. We let the workers to make at lease two answers with both knowledge and persona among the six answers of the dialog, and mark the sources they've used.}
\end{figure}

\clearpage

\subsection{Grounding Quality Evaluation Pages}
\begin{figure}[h!]
	\centering
	\includegraphics[width=0.85\linewidth]{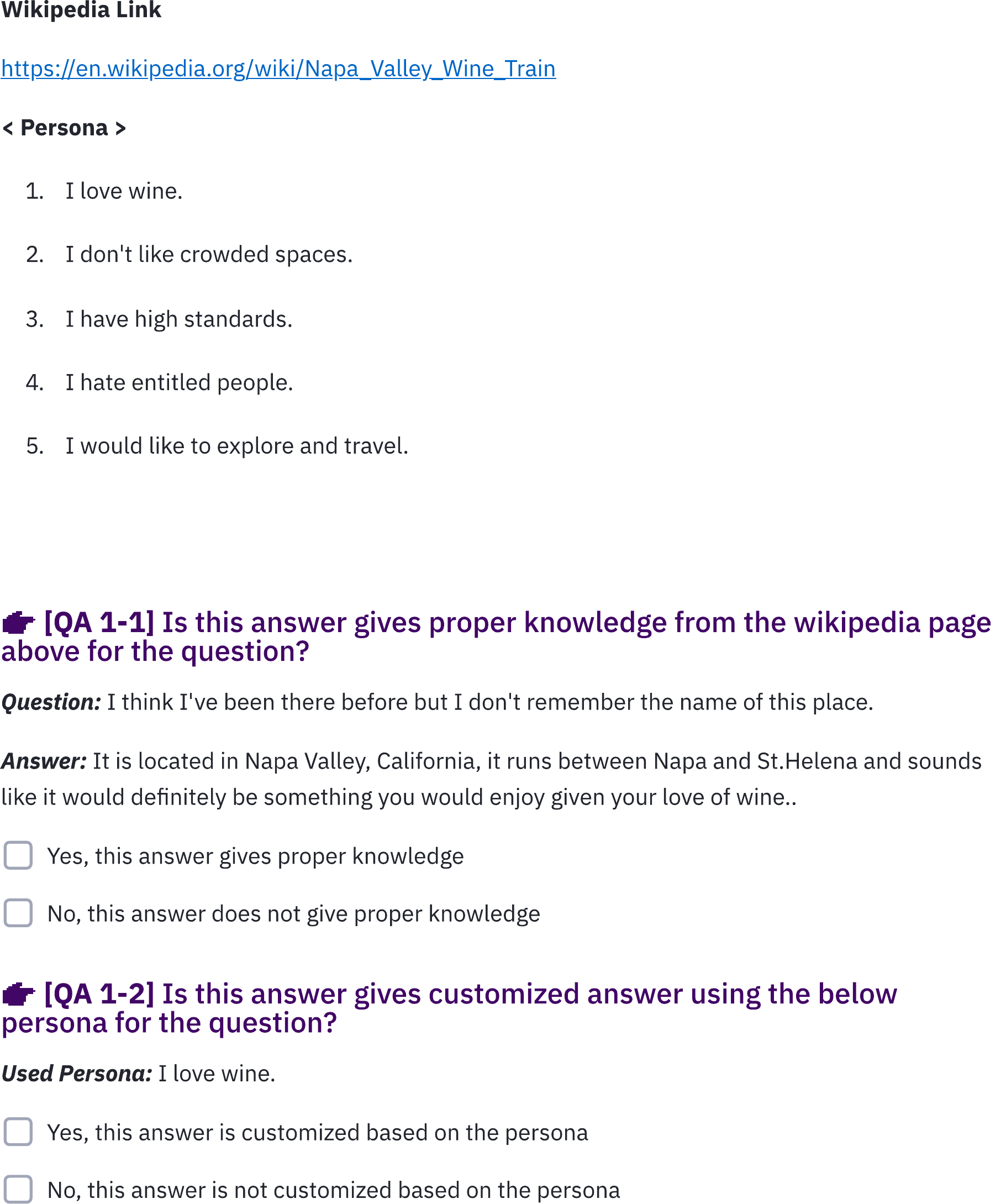}
	\includegraphics[width=0.85\linewidth]{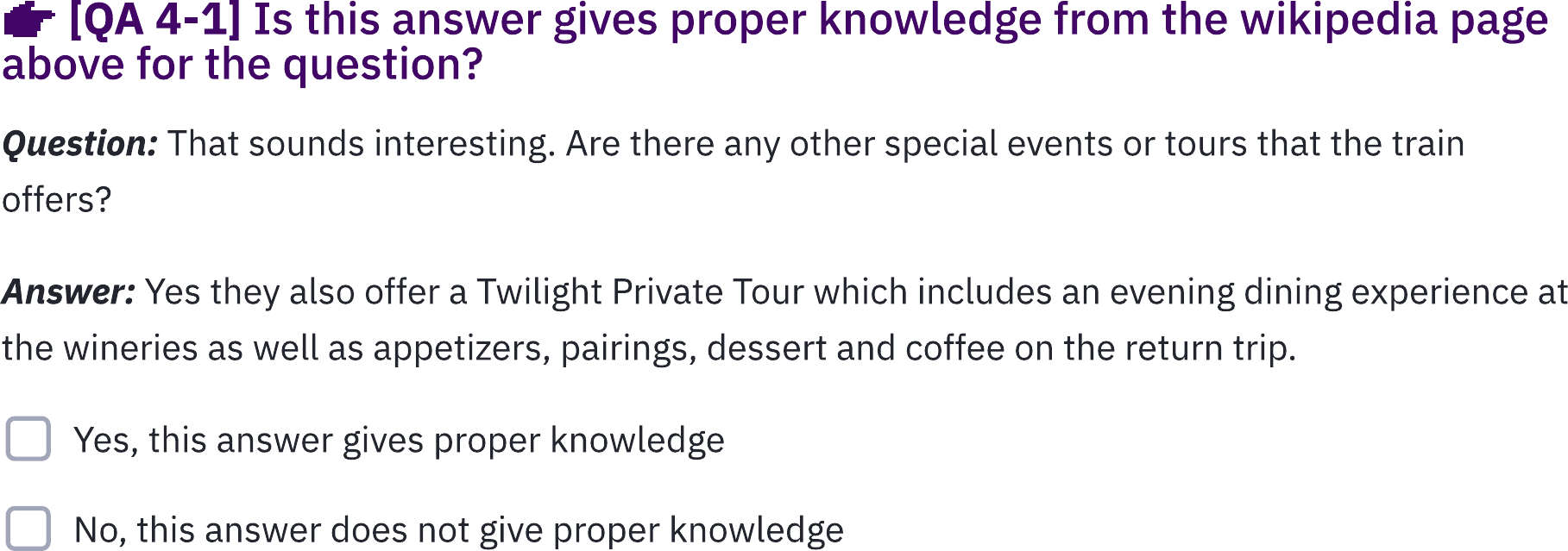}
\caption{Pages on grounding quality evaluation. Human evaluators checks whether the utterances in the data are well-grounded, given the persona and the Wikipedia page link.}
\end{figure}

\clearpage
\subsection{Human Evaluation Pages}
\begin{figure}[h]
	\centering
	\includegraphics[width=0.95\linewidth]{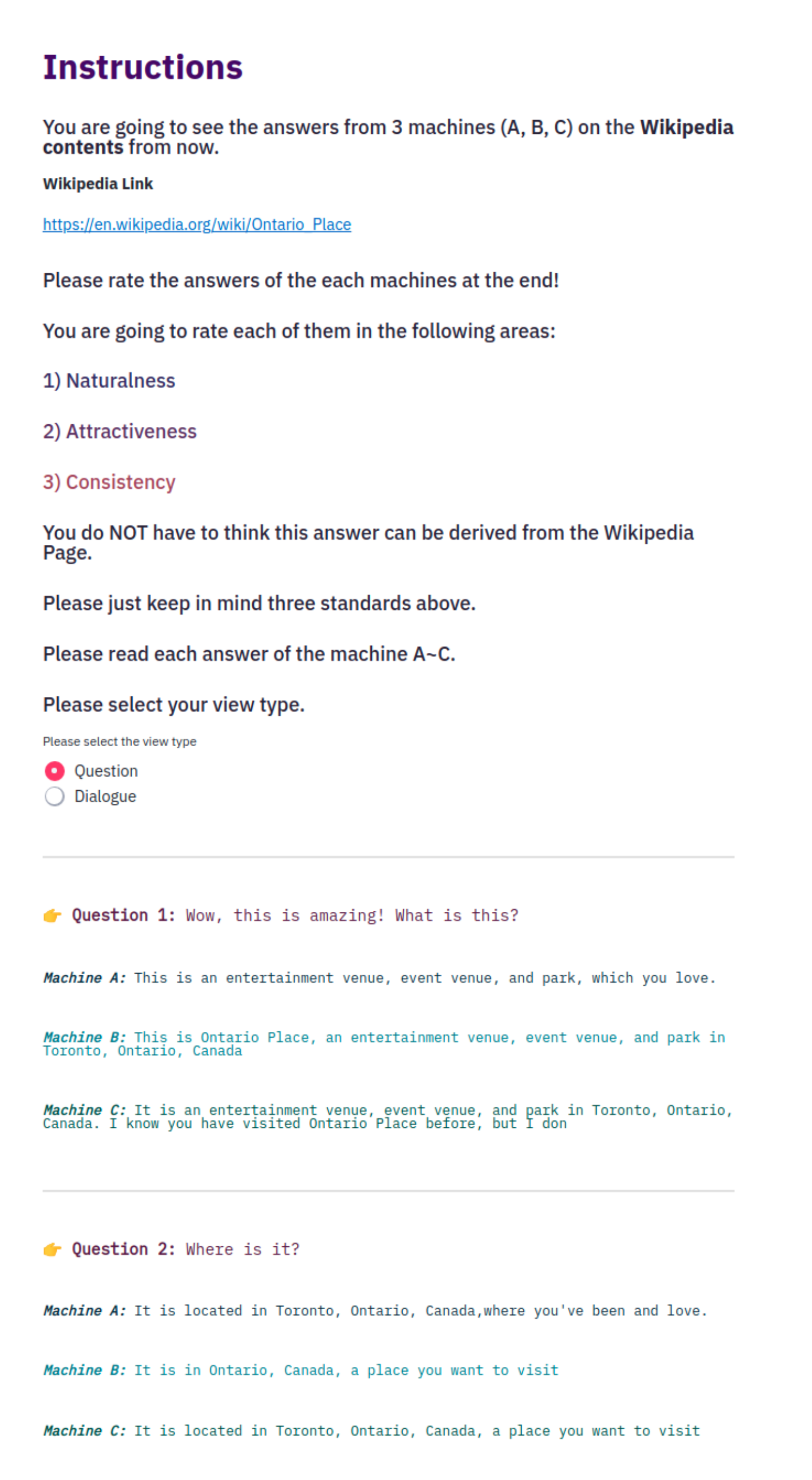}
\caption{Introduction page on utterance quality evaluation.}
\end{figure}

\begin{figure}[h]
	\centering
	\includegraphics[width=0.85\linewidth]{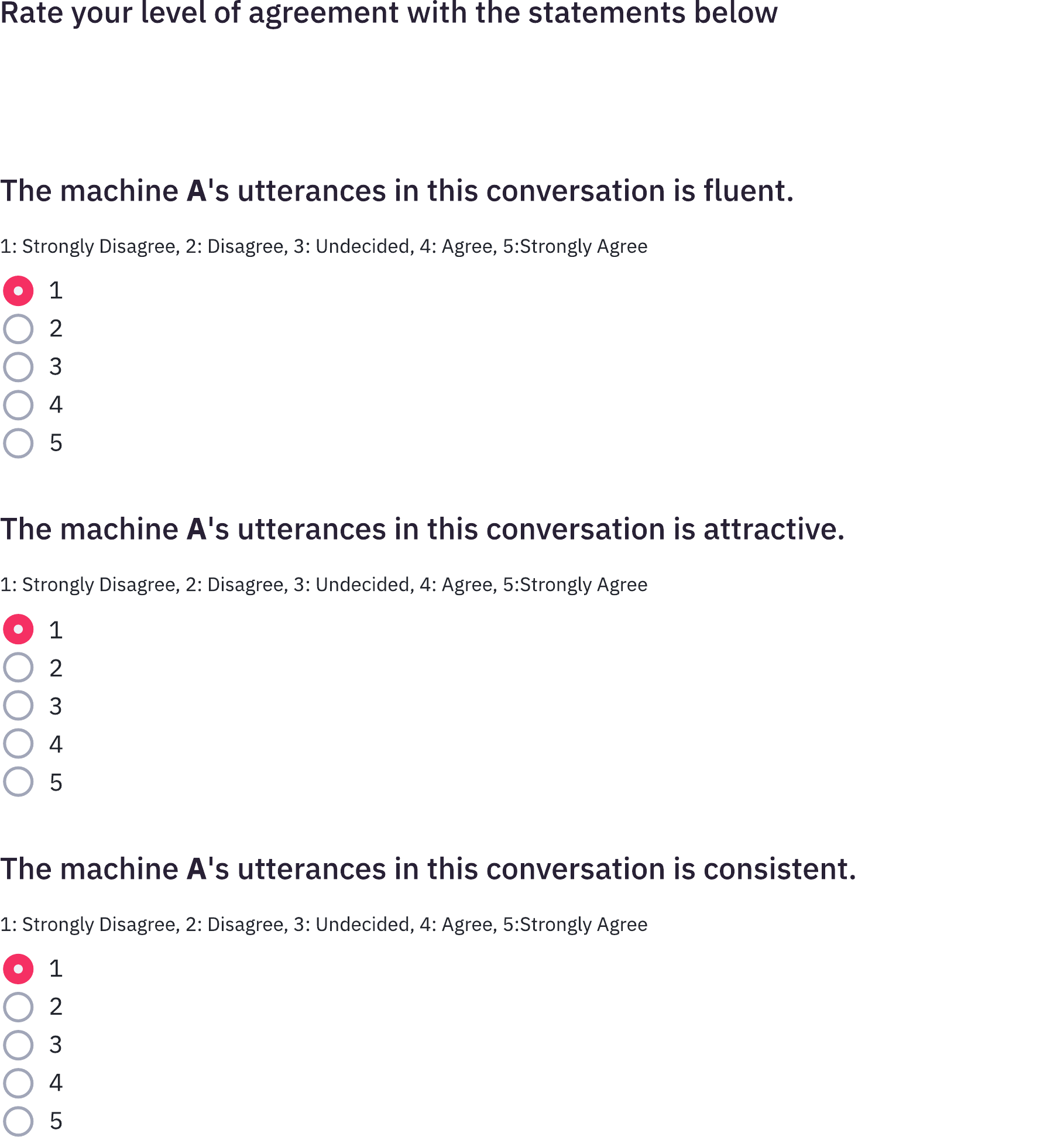}
\caption{Rating page on utterance quality evaluation. Human evaluators rate the utterances on fluency, engagement(attractiveness) and consistency. }
\end{figure}

\vfill\eject
\onecolumn
\subsection{Dataset Exploration}
\begin{figure*}[!h]
	\centering
	\includegraphics[width=0.80\linewidth]{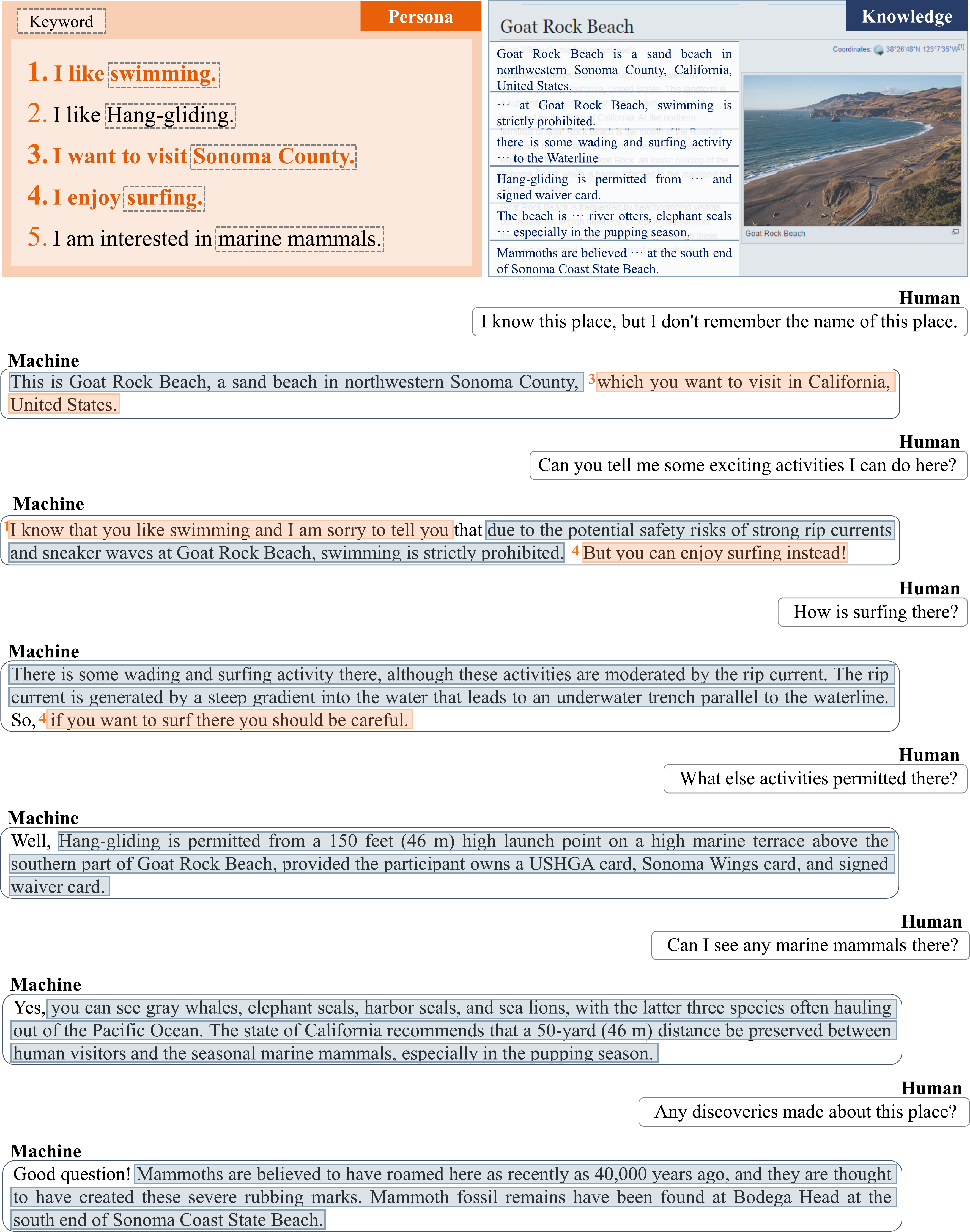}\\
\caption{An example dialog 1. The workers firstly make persona sentences given the Wikipedia page link. With the persona sentences they made, they make a dialog between the human and the machine. The human asks questions regarding the landmark, and the machine gives answers considering both the knowledge and the persona of the user.}
\end{figure*}

\begin{figure*}[h]
	\centering
	\includegraphics[width=0.80\linewidth]{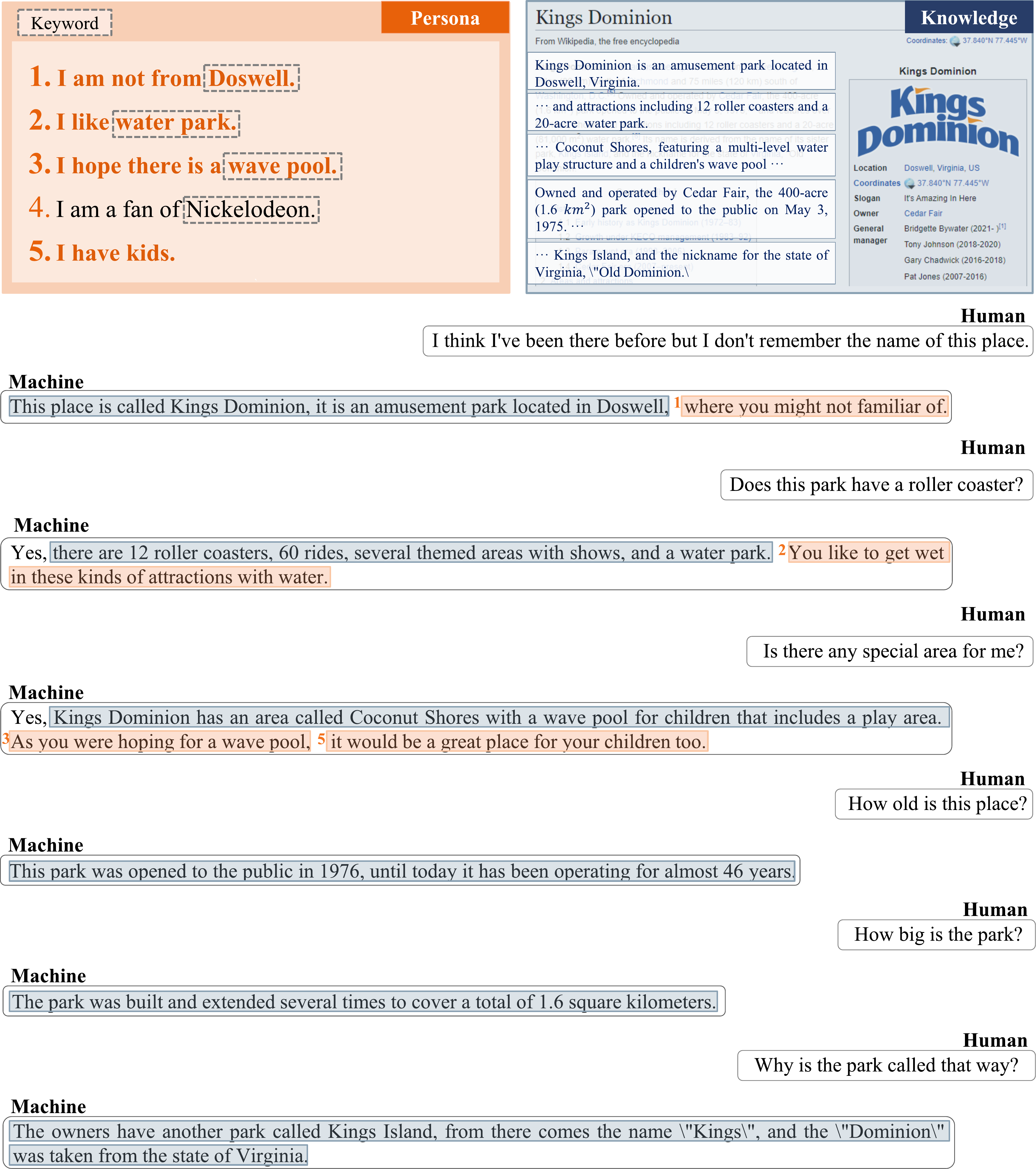}\\
\caption{An example dialog 2. The machine gives a response with the retrieved knowledge and the persona of the user. The number in front of the orange colored utterances indicates the persona number of the user. The utterances in blue boxes are made with the Wikipedia knowledge, which is shown in the knowledge box.}
\end{figure*}

\begin{figure*}[h]
	\centering
	\includegraphics[width=0.85\linewidth]{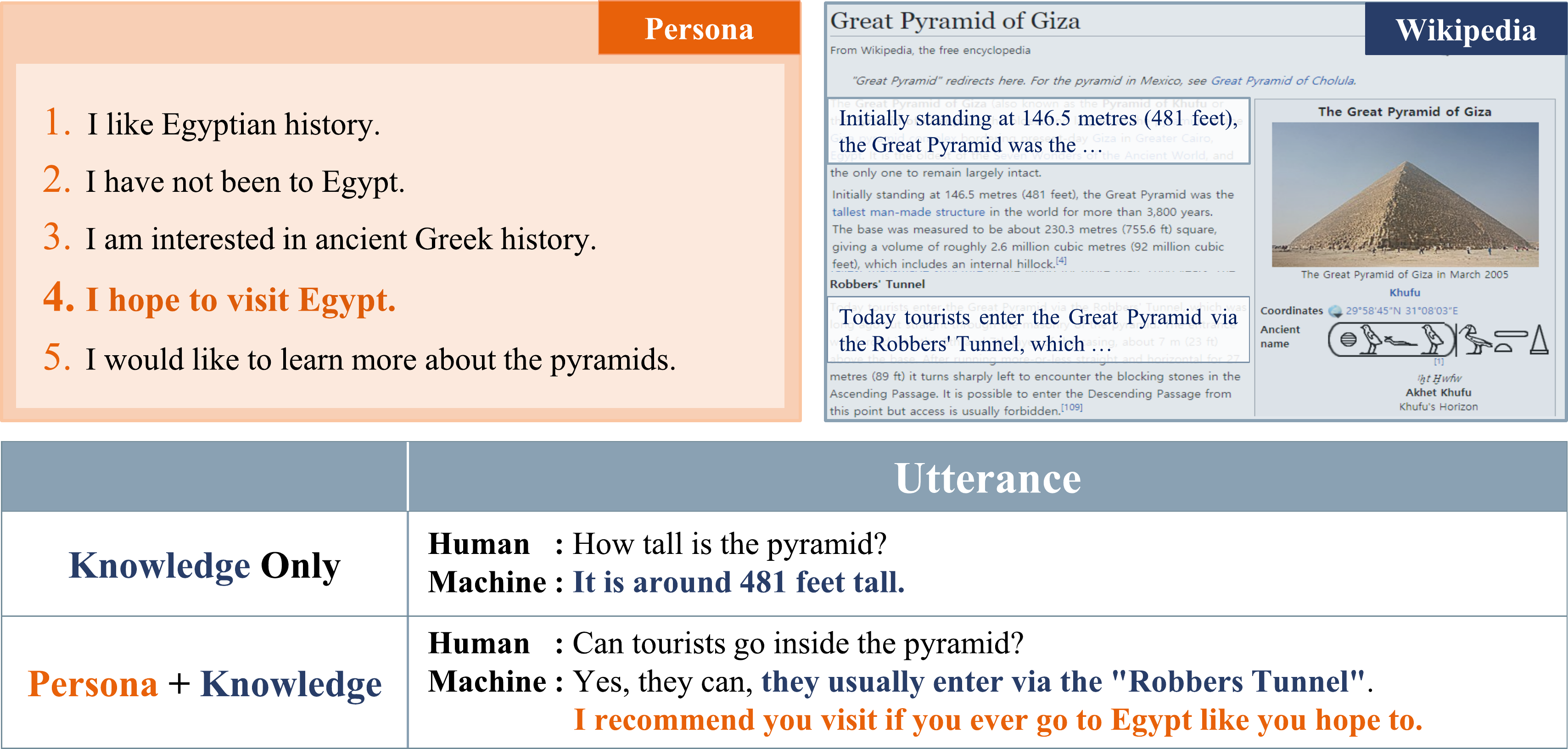}\\
	\includegraphics[width=0.85\linewidth]{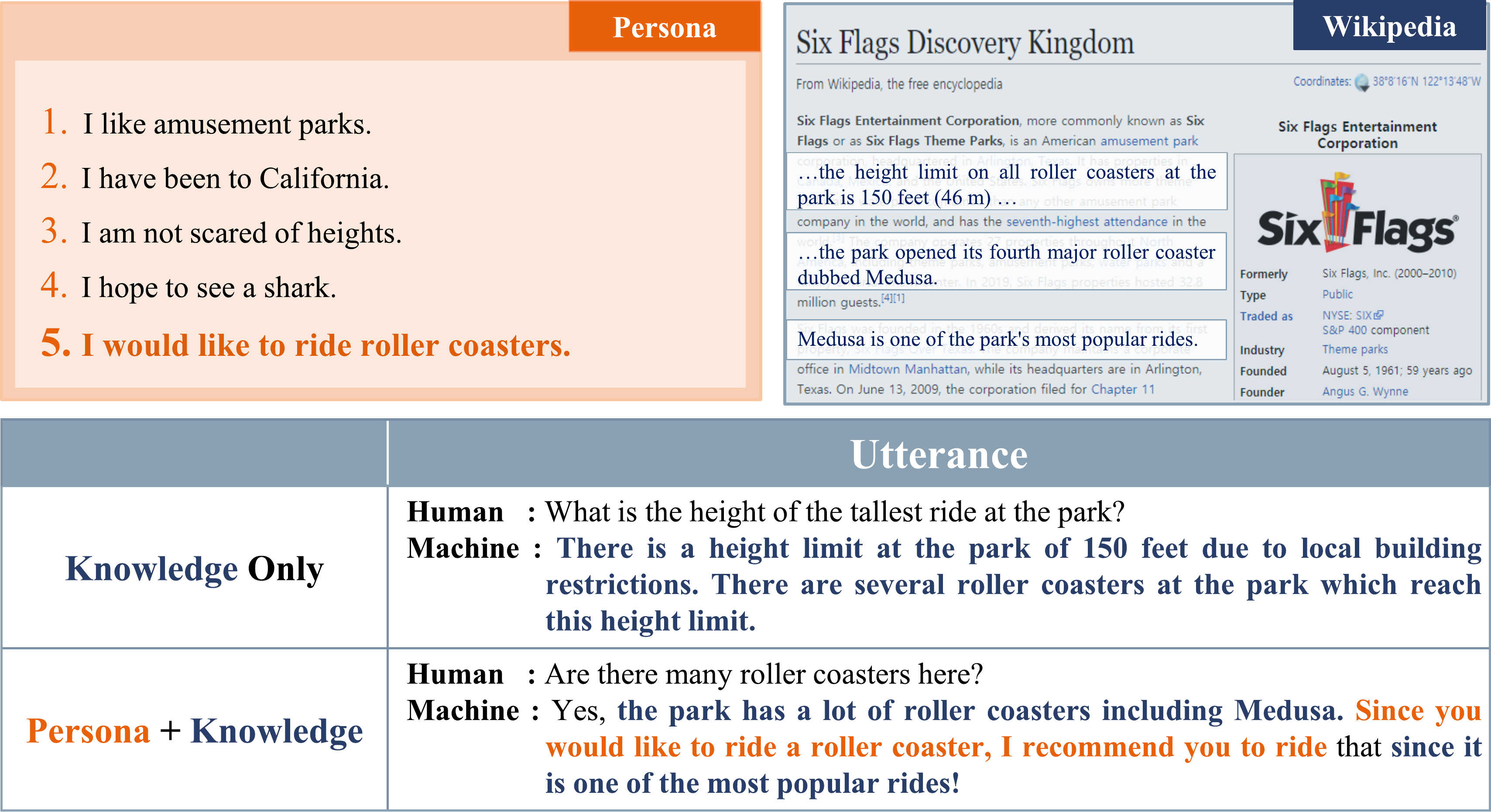}\\
\caption{Answer Types: Answers with Knowledge-Only and Persona-Knowledge. The answers of the machine utilize the knowledge source or both of the persona and the knowledge source to answer the question of the human. The sentences in blue are from the Wikipedia page, and the sentences in orange are from the user's persona. The utterances with both persona and knowledge are the customized answers for the user.}
\end{figure*}

\onecolumn
\begin{figure*}[h]
	\centering
	\includegraphics[width=0.85\linewidth]{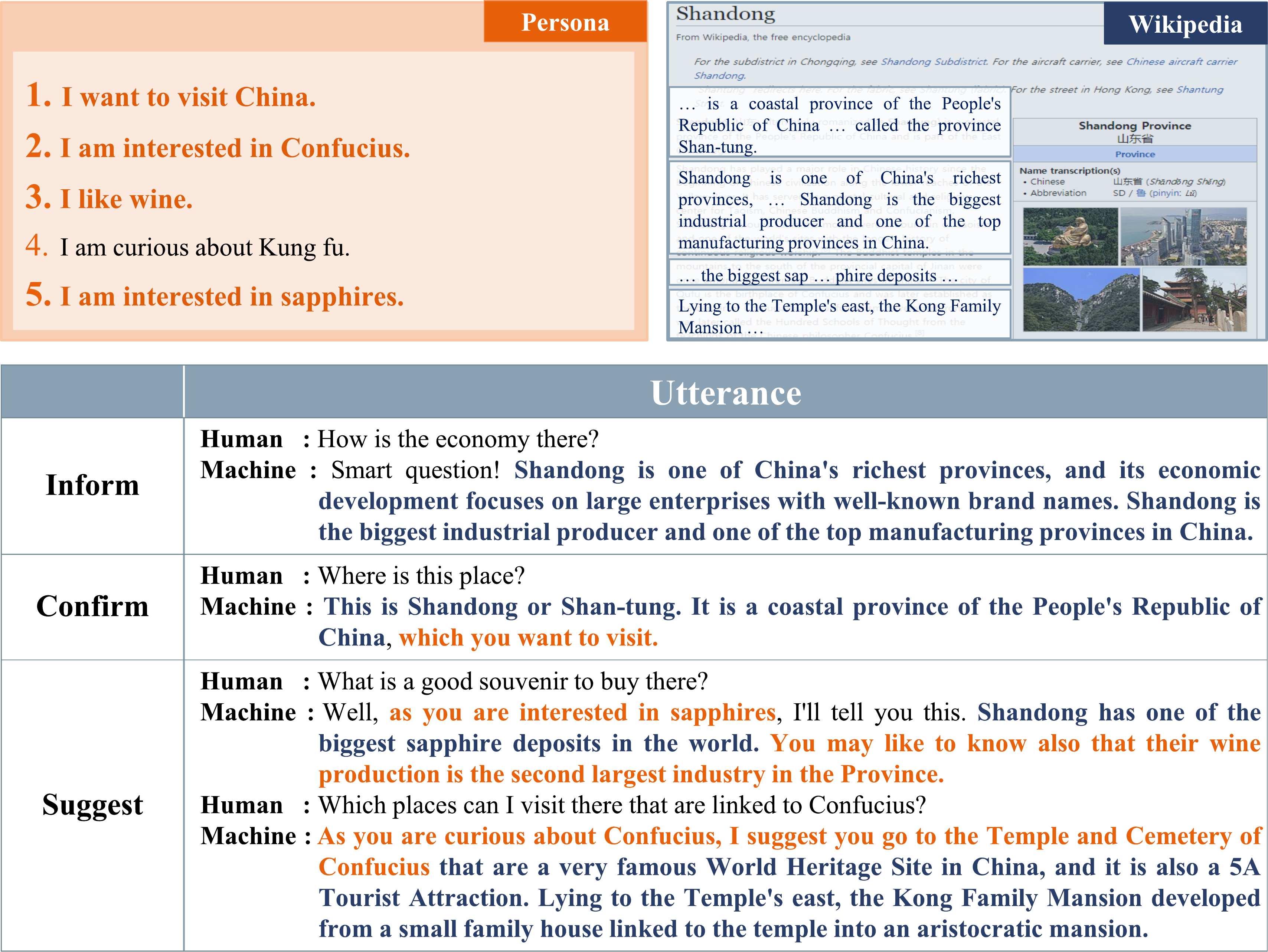}\\
	\includegraphics[width=0.85\linewidth]{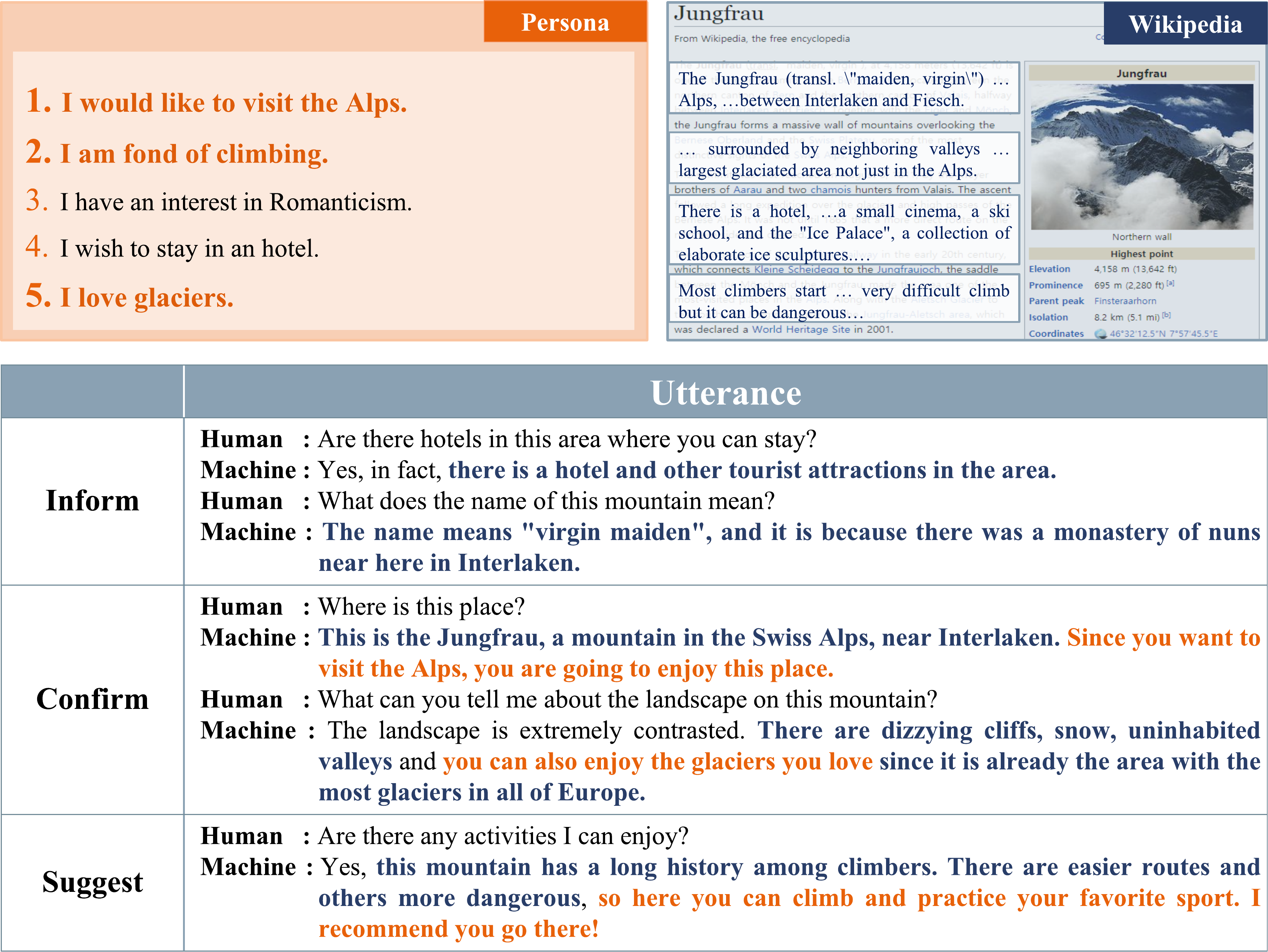}\\
 \caption{Answer Types: Inform, Confirm and Suggest The answers of the machine are categorized into three types, which are \emph{Inform}, \emph{Confirm} and \emph{Suggest}. \emph{Inform} type of answers just give proper knowledge to the user, and \emph{Confirm} type of answers show intimacy based on the user's persona. \emph{Suggest} type of answers recommend certain thing to the users considering their persona.}
\end{figure*}
\clearpage

\subsection{Conversations between human and our model}
\begin{table}[h!]
\centering
\begin{tabular}{|p{13.0cm}|}
\hline
\textbf{Persona} \\
\hline
1. I am a resident of United States. \\
2. I like to visit historic places. \\
3. I am a bird lover. \\
4. I hope to see fossils. \\
5. I love fishing and camping. \\
\hline
\textbf{Landmark} \\
\hline
https://en.wikipedia.org/wiki/Worlds\_End\_State\_Park\\
\hline
\textbf{Dialog} \\
\hline
\textit{Human} : Wow, this is amazing! What is this? \\
\textit{Model} : This is Worlds End State Park, a 780-acre (316 ha) Pennsylvania state park in Sullivan \\ 
\textit{Human} : Good! What can I see here? \\
\textit{Model} : Since you like wildlife, you will be interested to know that the park has over 200 species of birds  \\ 
\textit{Human} : Nice to hear! What can I do here for recreations? \\
\textit{Model} : Yes, you can go hiking, camping, rafting, snowmobiling, cross-country skiing \\ 
\textit{Human} : The name sounds interesting! What is the reason for the park name? \\
\textit{Model} : The name has caused some confusion and controversy over the years.  Although it was founded as Worlds End \\ 
\textit{Human} : Interesting! Does the park has any special feature? \\
\textit{Model} : Yes, the park is one of the most popular tourist destination in the region and is known to attract  \\ 
\textit{Human} : Can we see animals here? \\
\textit{Model} : Yes, there are deer, bears, wild turkeys, red and gray squirrels, and bob  \\ 
\hline
\end{tabular}

\vspace{2mm}

\begin{tabular}{|p{13.0cm}|}
\hline
\textbf{Persona} \\
\hline
1. I'm looking for an educational trip. \\
2. I'm interested in art. \\
3. I'm interested in Indian Americans. \\
4. I'd like to visit Seattle. \\
5. I want to attend an event. \\
\hline
\textbf{Landmark} \\
\hline
https://en.wikipedia.org/wiki/Daybreak\_Star\_Cultural\_Center \\
\hline
\textbf{Dialog} \\
\hline
\textit{Human} : Wow, this is amazing! What is this? \\
\textit{Model} : This is the Daybreak Star Indian Cultural Center in Seattle, Washington, which you want to visit \\ 
\textit{Human} : Who founded Daybreak Star Indian Cultural Center? \\
\textit{Model} : The center developed from activism by Bernie Whitebear and other Native Americans, who staged a generally successful self \\ 
\textit{Human} : What is the architectural style of Daybreak Star Indian Cultural Center? \\
\textit{Model} : Daybreak Star Indian Cultural Center was built in the Renaissance Revival style of architecture. \\ 
\textit{Human} : How large is Daybreak Star Indian Cultural Center? \\
\textit{Model} : The center has an area of 81,000 square meters. \\ 
\textit{Human} : What is the event I can attend at Daybreak Star Indian Cultural Center? \\
\textit{Model} : You can go for pow wows, which is something you've never done. \\ 
\textit{Human} : What is displayed at Daybreak Star Indian Cultural Center? \\
\textit{Model} : You will be glad to know that the center has a permanent art collection that includes a variety of large \\ 
\hline
\end{tabular}

\caption{Examples of selected conversations between human and the model. GPT-2$_{Medium}$ trained with PG and KG, which is a large-scale model of GPT-2$_{Small}$.}
\label{table:selected_conversation}
\end{table}

\end{document}